\documentclass{article}

\usepackage{microtype}
\usepackage{graphicx}
\usepackage{subfigure}
\usepackage{booktabs} % for professional tables

\usepackage{hyperref}

\usepackage[accepted]{icml2025}

\usepackage{amsmath}
\usepackage{amssymb}
\usepackage{mathtools}
\usepackage{amsthm}

\usepackage[capitalize,noabbrev]{cleveref}

\theoremstyle{plain}

\theoremstyle{definition}

\theoremstyle{remark}

\usepackage[textsize=tiny]{todonotes}

\newcommand{\tname}[1]{ID$^2$}
\newcommand{\dname}[1]{OriPID}
\newcommand{\dsname}[1]{TIPS-I2V}

\usepackage{amsmath}       
\usepackage{array}       
\usepackage{amsthm}
\usepackage{bm}

\usepackage{xcolor}
\usepackage{tcolorbox}
\tcbuselibrary{breakable}
\definecolor{lightroyalblue}{HTML}{F6F8FD}
\definecolor{mygray}{gray}{.9}
\definecolor{ggray}{RGB}{127,127,127}
\definecolor{reda}{RGB}{192,0,0}
\definecolor{redb}{RGB}{217,148,143}
\definecolor{myyellow}{RGB}{190,144,0}
\definecolor{mygreen}{RGB}{80,100,40}
\definecolor{myblue}{RGB}{30,90,100}
\definecolor{mygray1}{RGB}{245,245,245}
\usepackage{tabularx}
\newcolumntype{Y}{>{\centering\arraybackslash}X}
\usepackage{multirow}
\usepackage{caption}
\usepackage{graphicx}
\usepackage{tabularx}
\usepackage{colortbl}
\usepackage{fontawesome}
\usepackage{footmisc}
\definecolor{palegreen}{HTML}{FCF8FD}

\definecolor{pp}{HTML}{964A6B}
\definecolor{bb}{HTML}{3476B9}
\makeatletter  % as a whole with thickhline
\newcommand{\thickhline}{%
	\noalign {\ifnum 0=`}\fi \hrule height 1pt
	\futurelet \reserved@a \@xhline
}
\newtcolorbox{abox}{colback=lightroyalblue,colframe=black,boxrule=0.5pt}
\usepackage{mdframed}
\icmltitlerunning{Origin Identification for Text-Guided Image-to-Image Diffusion Models}

\begin{document}

\twocolumn[

{%
\renewcommand\twocolumn[1][]{#1}%
\icmltitle{Origin Identification for Text-Guided Image-to-Image Diffusion Models}

%\vspace{-2mm}




\begin{icmlauthorlist}
\icmlauthor{Wenhao Wang}{aaa}
\icmlauthor{Yifan Sun}{ccc}
\icmlauthor{Zongxin Yang}{ddd}
\icmlauthor{Zhentao Tan}{bbb}
\icmlauthor{Zhengdong Hu}{aaa}
\icmlauthor{Yi Yang}{ccc}
\end{icmlauthorlist}

\icmlaffiliation{aaa}{University of Technology Sydney}
\icmlaffiliation{bbb}{Peking University}
\icmlaffiliation{ddd}{Harvard University}
\icmlaffiliation{ccc}{Zhejiang University}

\icmlcorrespondingauthor{Yi Yang}{yangyics@zju.edu.cn}
%\icmlcorrespondingauthor{Firstname2 Lastname2}{first2.last2@www.uk}

%\icmlkeywords{Machine Learning, ICML}

%\vskip 0.3in

\begin{center}
    \centering
    \vspace{1.em}
    \includegraphics[width=1\textwidth]{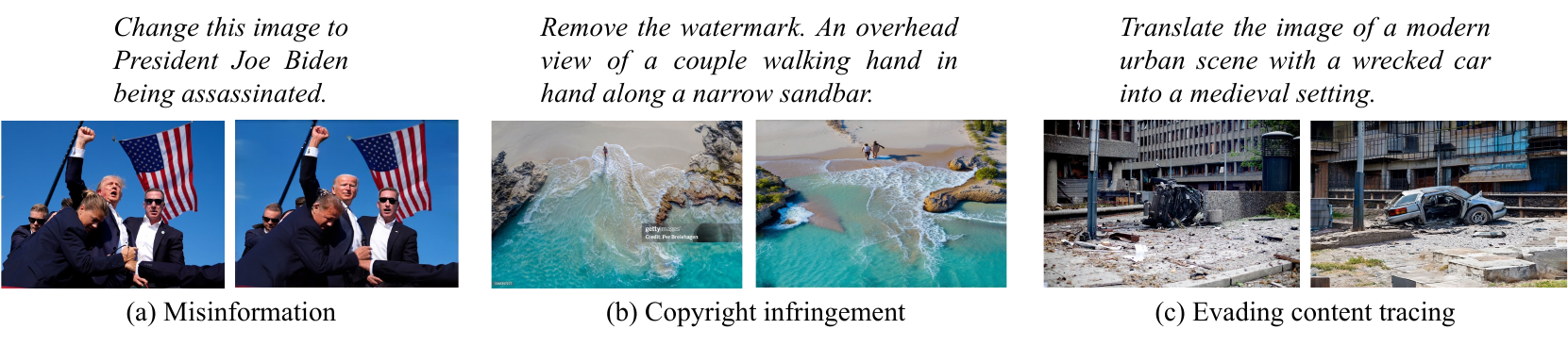}
    \vspace{-2em}
    \captionof{figure}{The illustration for \textbf{misusing} text-guided image-to-image diffusion models in several scenarios: \textit{misinformation}, \textit{copyright infringement}, and \textit{evading content tracing}. Specifically: \textbf{(a)} An altered image originally showing Donald Trump post-assassination is edited to depict Joe Biden instead; \textbf{(b)} The removal of a watermark from a copyrighted beach image, followed by modifications, could assist in escaping copyright checks; \textbf{(c)} An image of a Norwegian government building after an explosion is altered to bypass restrictions, which limit the spread of disturbing images.}
    \label{Fig: misuse}
    %\vspace{1mm}
\vspace{2mm}
\end{center}
}

]
\printAffiliationsAndNotice{}

\begin{abstract}
%\vspace{-0.mm}
Text-guided image-to-image diffusion models excel in translating images based on textual prompts, allowing for precise and creative visual modifications. However, such a powerful technique can be misused for \textit{spreading misinformation}, \textit{infringing on copyrights}, and \textit{evading content tracing}. This motivates us to introduce the task of origin \textbf{ID}entification for text-guided \textbf{I}mage-to-image \textbf{D}iffusion models (\textbf{ID$\mathbf{^2}$}), aiming to retrieve the original image of a given translated query. A straightforward solution to ID$^2$ involves training a specialized deep embedding model to extract and compare features from both query and reference images. However, due to \textit{visual discrepancy} across generations produced by different diffusion models, this similarity-based approach fails when training on images from one model and testing on those from another, limiting its effectiveness in real-world applications. 
To solve this challenge of the proposed \tname~ task, we contribute the first dataset and a theoretically guaranteed method, both emphasizing generalizability.
The curated dataset, \textbf{\dname~}, contains abundant \textbf{Ori}gins and guided \textbf{P}rompts, which can be used to train and test potential \textbf{ID}entification models across various diffusion models.
In the method section, we first prove the \textit{existence} of a linear transformation that minimizes the distance between the pre-trained Variational Autoencoder embeddings of generated samples and their origins.
Subsequently, it is demonstrated that such a simple linear transformation can be \textit{generalized} across different diffusion models.
Experimental results show that the proposed method achieves satisfying generalization performance, significantly surpassing similarity-based methods ($+31.6\%$ mAP), even those with generalization designs.
The project is available at \url{https://id2icml.github.io}.
\vspace{-3mm}
\end{abstract}

\section{Introduction}

\begin{figure*}[t]
    \centering
    \includegraphics[width=0.98\textwidth]{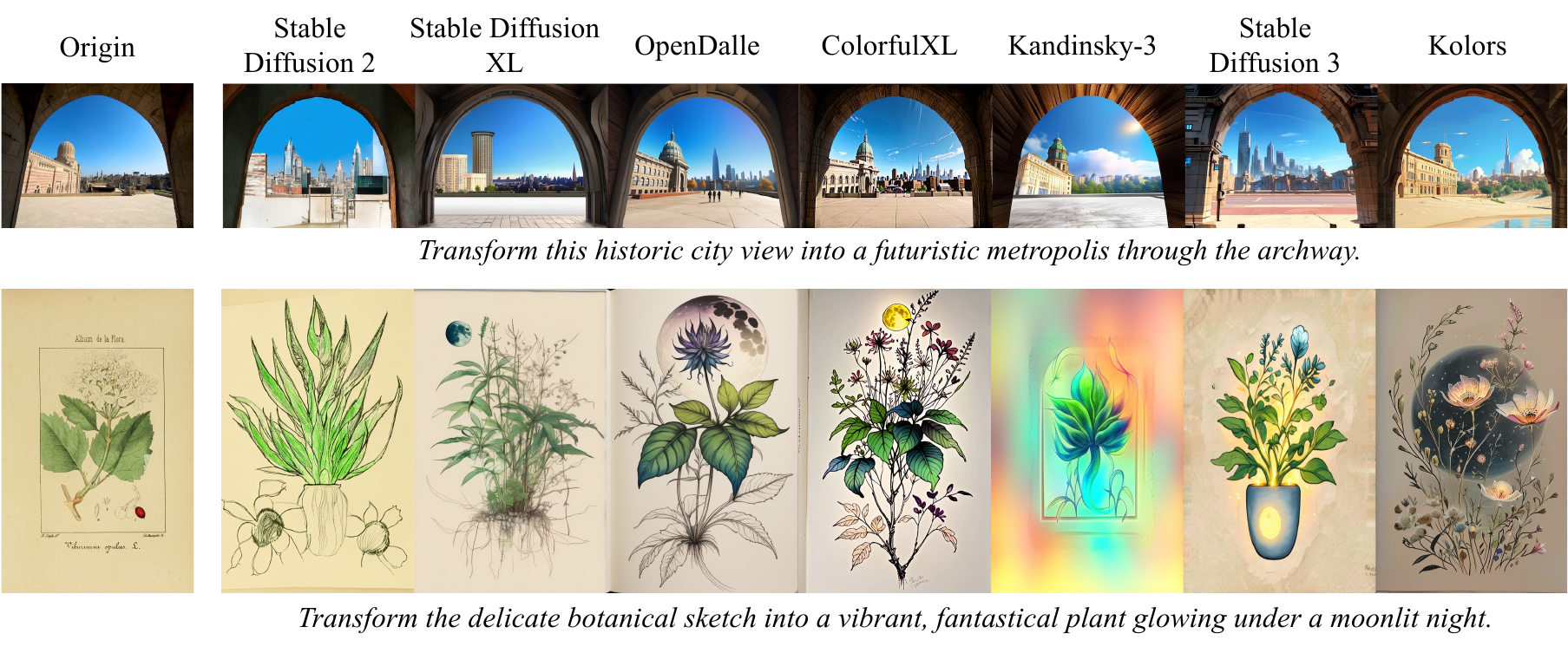}
    \vspace{-3mm}  
    \caption{The demonstration for \textit{visual discrepancy} between generated images by different diffusion models. The images generated by various models exhibit distinctive visual features such as realistic textures, complex architectures, life-like details, vibrant colors, abstract expression, magical ambiance, and photorealistic elements.} 
    \label{Fig: VD}
   \vspace{-4mm} 
\end{figure*}

Text-guided image-to-image diffusion models are notable for their ability to transform images based on textual descriptions, allowing for detailed and highly customizable modification. While they are increasingly used in creative industries for tasks such as 
digital art re-creation, customizing visual content, and personalized virtual try-ons, there are growing security concerns associated with their \textbf{misuse}.
As illustrated in Fig. \ref{Fig: misuse}, for instance, they could be misused for \textit{misinformation}, \textit{copyright infringement}, and \textit{evading content tracing}. To help combat these misuses, this paper introduces the task of origin \textbf{ID}entification for text-guided \textbf{I}mage-to-image \textbf{D}iffusion models (\textbf{ID$\mathbf{^2}$}), which aims to identify the original image of a generated query from a large-scale reference set. When the origin is identified, subsequent compensations include deploying factual corrections for misinformation, enforcing copyright compliance, and keeping the tracing of target content.\par 

A straightforward solution for the proposed \tname~ task is to employ a similarity-based retrieval approach. Specifically, this approach \textbf{(1)} fine-tunes a pre-trained network by minimizing the distances between generated images and their origins, and \textbf{(2)} uses the trained network to extract and compare feature vectors from the queries and references. 
However, this approach is \textbf{impractical} in real-world scenarios.
This is because: for most current popular diffusion models, such as Stable Diffusion 2 \citep{Rombach_2022_CVPR}, Stable Diffusion XL \citep{podell2024sdxl}, OpenDalle \citep{dataautogpt3}, ColorfulXL \citep{ColorfulXL}, Kandinsky-3 \citep{arkhipkin2023kandinsky}, Stable Diffusion 3 \citep{esser2024scaling}, and Kolors \citep{kolors}, in a training-free manner, text-guided image-to-image translation can be easily achieved by using an input image with added noise as the starting point (instead of starting from randomly distributed noise). Further, as shown in Fig. \ref{Fig: VD}, there exists a \textit{visual discrepancy} across images generated by different diffusion models, \textit{i.e.}, different diffusion models exhibit distinct visual features. An experimental evidence for such discrepancy is that we can train a lightweight classification model, such as Swin-S \citep{liu2021swin}, to achieve a top-1 accuracy of \textbf{95.9\%} when classifying images generated by these seven diffusion models.
The visual discrepancy presents an inherent challenge of our \tname~, \textit{i.e.}, \textit{the approach mentioned above fails when trained on images generated by one diffusion model and tested on queries from another}. For instance, when trained on images generated by Stable Diffusion 2, this approach achieves a \textbf{87.1\%} mAP on queries from Stable Diffusion 2, while only achieving a \textbf{30.5\%} mAP on ColorfulXL. %In conclusion, the visual discrepancy hinders similarity-based retrieval approaches from being generalizable across different diffusion models, limiting their real-world usefulness.

To address the generalization challenge in the proposed task, our efforts focus primarily on \textit{constructing the first \tname~ dataset} and \textit{proposing a theoretically guaranteed method}.

$\bullet$ \textbf{A new dataset emphasizing generalization.} To verify the generalizability, we construct the first \tname~ dataset, \textbf{\dname~}, which includes abundant \textbf{Ori}gins with guided \textbf{P}rompts for training and testing potential \textbf{ID}entification models. Specifically, the \textit{training} set contains $100,000$ origins. For each origin, we use GPT-4o \citep{openai2024gpt4o} to generate $20$ different prompts, each of which implies a plausible translation direction. By inputting these origins and prompts into Stable Diffusion 2, we generate $2,000,000$ training images. For \textit{testing}, we randomly select $5,000$ images as origins from a reference set containing $1,000,000$ images, and ask GPT-4o to generate a guided prompt for each origin. Subsequently, we generate $5,000$ queries using the origins, corresponding prompts, and each of the following models: Stable Diffusion 2, Stable Diffusion XL, OpenDalle, ColorfulXL, Kandinsky-3, Stable Diffusion 3, and Kolors. The design of using different diffusion models to generate training images and queries is particularly practical because, in the real world, where numerous diffusion models are publicly available, we cannot predict which ones might be misused. %In short, verifying the generalizability of trained identification models is necessary.\par 
%, and (3) implements cosine similarity to compare the feature vectors of the queries against those of the references

$\bullet$ \textbf{A simple, generalizable, and theoretically guaranteed solution.}
To solve the generalization problem, we first theoretically prove that, after specific linear transformations, the embeddings of an original image and its translation, encoded by the diffusion model’s Variational Autoencoder (VAE), will be sufficiently close. This suggests that we can use these linearly transformed query embeddings to match against the reference embeddings. Furthermore, we demonstrate that these kinds of feature vectors are generalizable across diffusion models. Specifically, by using a trained linear transformation and the encoder of VAE from one diffusion model, we can also effectively embed the generated images from another diffusion model, even if their VAEs have different parameters or architectures (see Section \ref{Sec: VAE} for more details). The effectiveness means the similar performance of origin identification for both diffusion models. Finally, we implement this theory (obtain the expected linear transformation) by gradient descending a metric learning loss and experimentally show the effectiveness and generalizability of the proposed solution.

In summary, we make the following contributions:
\vspace{-2mm}
\begin{enumerate}
\item This paper proposes a novel task, origin identification for text-guided image-to-image diffusion models (\tname~), which aims to identify the origin of a generated query. This task tries to alleviate an important and timely security concern, \textit{i.e.}, the misuse of text-guided image-to-image diffusion models. To support this task, we build the first \tname~ dataset.

\item We highlight an inherent challenge of \tname~, \textit{i.e.}, the existing visual discrepancy prevents similarity-based methods from generalizing to queries from unknown diffusion models. Therefore, we propose a simple but generalizable method by utilizing linear-transformed embeddings encoded by the VAE. Theoretically, we prove the existence and generalizability of the required linear transformation.

\item Extensive experimental results show (1) the challenge of the proposed \tname~ task: all pre-trained deep embedding models, fine-tuned similarity-based methods, and specialized domain generalization methods fail to achieve satisfying performance; and (2) the effectiveness of our proposed method: it achieves $88.8\%$, $81.5\%$, $87.3\%$, $89.3\%$, $85.7\%$, $85.7\%$, and $90.3\%$ mAP, respectively, for Stable Diffusion 2, Stable Diffusion XL, OpenDalle, ColorfulXL, Kandinsky-3, Stable Diffusion 3, and Kolors.
\end{enumerate}
\vspace{-2mm}

\section{Related Works}
%Diffusion models have become a transformative class of generative models, utilizing iterative noise-based processes to excel in tasks such as image synthesis, inpainting, and text-to-image generation. By progressively denoising data, these models can reconstruct highly detailed images, offering flexibility and precision in creative applications. 
\textbf{Text-guided Image-to-image Diffusion Models.}
Recent diffusion models, including Stable Diffusion 2 \citep{Rombach_2022_CVPR}, Stable Diffusion XL \citep{podell2024sdxl}, OpenDalle \citep{dataautogpt3}, ColorfulXL \citep{ColorfulXL}, Kandinsky-3 \citep{arkhipkin2023kandinsky}, Stable Diffusion 3 \citep{esser2024scaling}, and Kolors \citep{kolors}, have brought significant improvements in visual generation. This paper considers using these popular models for text-guided image-to-image translation as SDEdit \citep{mengsdedit}, which is common and cost-effective in the real world. 
We also note that certain methods, such as InstructPix2Pix \citep{brooks2023instructpix2pix}, IP-Adapter \citep{ye2023ip-adapter}, EDICT \citep{wallace2023edict}, and Plug-and-Play \citep{tumanyan2023plug}, perform this task in other paradigms. For a detailed discussion, please refer to Appendix (Section \ref{App: future}).

%\textbf{Text-guided Image-to-Image Translation.}
%In addition to the popular text-to-image capabilities of diffusion models, they can also creatively modify existing images based on textual prompts. In a training-free manner, this functionality can be easily achieved by using an input image with added noise as the starting point, instead of starting from randomly distributed noise. Most of current popular diffusion models, such as Stable Diffusion 2 \citep{Rombach_2022_CVPR}, Stable Diffusion XL \citep{podell2024sdxl}, OpenDalle \citep{dataautogpt3}, ColorfulXL \cite{ColorfulXL}, Kandinsky-3 \citep{arkhipkin2023kandinsky}, Stable Diffusion 3 \citep{esser2024scaling}, and Kolors \citep{kolors}, support this functionality, which enables users to seamlessly integrate new elements while preserving the overall structure of the original image. Instead of enhancing the quality or accuracy of text-guided image-to-image diffusion models, this paper focuses on identifying the origin of a given translated image. This approach offers benefits in combating misuse of these diffusion models, such as defeating misinformation, protecting copyright, and tracing content. We hope people become aware of the security risks brought by diffusion models while enjoying the creation of digital art.

\textbf{Security Issues with AI-Generated Content.}
Recently, generative models have gained significant attention due to their impressive capabilities. However, alongside their advancements, several security concerns have been identified. Prior research has explored various dimensions of these security issues. For instance, \citep{lin2024detecting} focuses on detecting AI-generated multimedia to prevent its associated societal disruption, and \citep{wang2024replication} explores replication problems in visual diffusion models. Additionally, \citep{fan2023trustworthiness} and \citep{chen2023challenges} explore the ethical implications and technical challenges in ensuring the integrity and trustworthiness of AI-generated content. 
In contrast, while our work also aims to help address the security issues, we specifically focus on a novel perspective: identifying the origin of a given translated image. 

\textbf{Image Copy Detection.}
The task most similar to our \tname~ is Image Copy Detection (ICD), which identifies whether a query \textit{replicates the content} of any reference. Various works focus on different aspects: PE-ICD \citep{wang2024pattern} and AnyPattern \citep{wang2024AnyPattern} build benchmarks and propose solutions emphasizing novel patterns in realistic scenarios; Active Image Indexing \citep{fernandez2022active} explores improving the robustness of ICD; and SSCD \citep{pizzi2022self} leverages self-supervised contrastive learning for ICD. Unlike ICD, which focuses on \textit{manually-designed transformations}, our \tname~ aims to find the origin of a query translated by the \textit{diffusion model with prompt-guidance}.

\section{Dataset}
\begin{figure*}[t]
    \centering
    \includegraphics[width=0.93\textwidth]{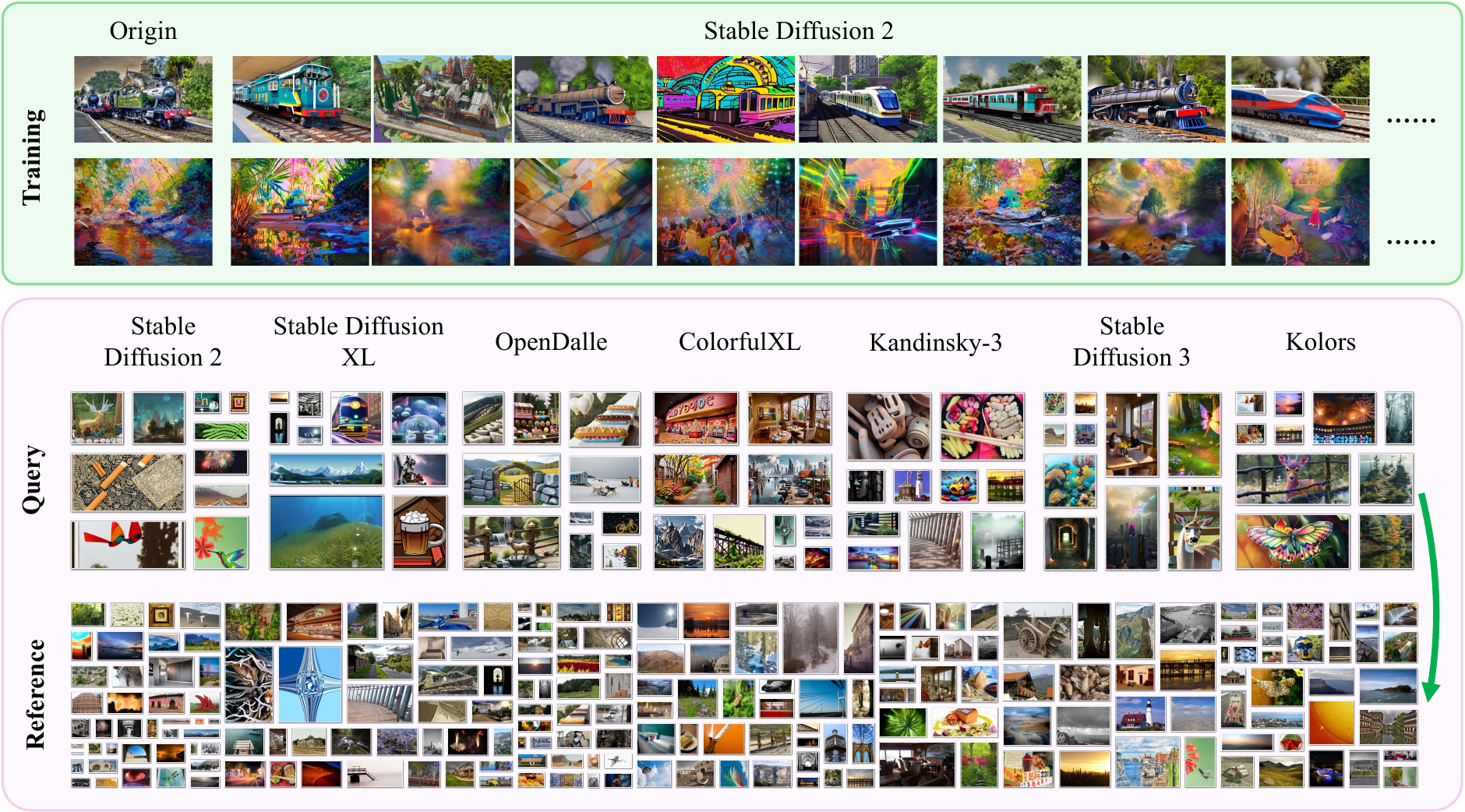}
    \vspace{-1mm}
    \caption{The images in our dataset, which is diverse and comprehensive. Specifically, it encompasses a variety of subjects commonly found in real-world scenarios where issues such as misinformation, copyright infringement, and content tracing evasion occur. For instance, our dataset includes images of nature, architecture, animals, planes, art, and indoor. Note that for simplicity, we omit the prompts here. Please refer to Appendix (Section \ref{App: prompt}) for examples of prompts and generations.} 
     \vspace{-4mm}
    \label{Fig: data}
\end{figure*}

To advance research in \tname~, this section introduces \dname~, the first dataset specifically designed for the proposed task. The source images in \dname~ are derived from the DISC21 dataset \citep{papakipos2022results}, which is a subset of the real-world multimedia dataset YFCC100M \citep{thomee2016yfcc100m}. As a result, \dname~ is diverse and comprehensive, covering a wide range of subjects related to real-world misinformation, copyright infringement, and content tracing evasion. An illustration of the proposed dataset is shown in Fig. \ref{Fig: data}. 

\textbf{Training Set.}
The training set comprises (1) $100,000$ origins randomly selected from the $1,000,000$ original images in DISC21, (2) $2,000,000$ guided prompts ($20$ for each origin) generated by GPT-4o (for details on how these prompts were generated, see Appendix (Section \ref{App: gpt})), and (3) $2,000,000$ images generated by inputting the origins and prompts into Stable Diffusion 2 \citep{Rombach_2022_CVPR}.

\textbf{Test Set.}
We design the test set with a focus on real-world/practical settings. On one hand, we use seven popular diffusion models, namely, Stable Diffusion 2 \citep{Rombach_2022_CVPR}, Stable Diffusion XL \citep{podell2024sdxl}, OpenDalle \citep{dataautogpt3}, ColorfulXL \citep{ColorfulXL}, Kandinsky-3 \citep{arkhipkin2023kandinsky}, Stable Diffusion 3 \citep{esser2024scaling}, and Kolors \citep{kolors}, to generate queries. This setting well simulates real-world scenarios where new diffusion models continuously appear, and we do not know which one is being misused. On the other hand, for each diffusion model, we generate $5,000$ queries to match $1,000,000$ references inherited from DISC21. This setting mimics the real world, where many distractors are not translated by any diffusion models.

\textbf{Scalability.}
Currently, we only use Stable Diffusion 2 to generate training images. However, our \dname~ can be easily scaled by incorporating more diffusion models for training, which may result in better generalizability. Furthermore, we only use $100,000$ origins and generate $20$ prompts for each origin. Researchers can scale up our dataset by using the entire $1,000,000$ original images and generating more prompts with the script in Appendix (Section \ref{App: gpt}).

\begin{table*}[t]
\caption{The $cos\left( \varphi \right)$ gained by compared Stable Diffusion 2 against different diffusion models. The experiments are repeated for \textbf{ten} times to calculate mean and standard deviation.} 
\vspace*{-2mm}
\small
  \begin{tabularx}{\hsize}{>{\centering\arraybackslash}p{1cm}|Y|Y|Y|Y|Y|Y}

    \hline
    $cos\left( \varphi \right) $&SDXL& OpenDalle & ColorfulXL& Kandinsky-3& SD3& Kolors\\ \hline
    \multirow{2 }{*}{SD2}& $0.995790$ ± $0.000037$ &  $0.996532$ ± $0.000016$& $0.998436$ ± $0.000015$& $0.999788$ ± $0.000009$& $0.993256$ ± $0.000035$& $0.991808$ ± $0.000042$\\
   \hline
  \end{tabularx}
  \label{Table: cos}
  \vspace*{-4mm}
\end{table*}
\vspace*{-2mm}
\section{Method}
To solve the proposed \tname~, we introduce a simple yet effective method, which is theoretically guaranteed and emphasizes generalizability. This section first presents two theorems regarding \textit{existence} and  \textit{generalizability}, respectively. \textbf{Existence} means that we can linearly transform the VAE embeddings of an origin and its translation such that their distance is close enough.
\textbf{Generalizability} means that the linear transformation trained on the images generated by one diffusion model can be effectively applied to another different diffusion model. Finally, we show how to train the required linear transformation in practice.% and an experimental evidence for generalizability.

\subsection{Existence}
\tcbset{colback=lightroyalblue, colframe=white, left=1mm, right=1mm, top=1mm, bottom=1mm}
\vspace*{-1mm}
\begin{tcolorbox}[breakable]
\textbf{Theorem 1.} \textit{Consider a well-trained diffusion model $\mathcal{F}_1$ with an encoder $\mathcal{E}_{1}$ from its VAE and its text-guided image-to-image functionability achieved by denoising noised images. There exists a linear transformation matrix $\mathbf{W}$, for any original image $o$ and its translation $g_1$, such that:}
\begin{equation}\label{Eq: T1}
\mathcal{E}_{1}(g_1) \cdot \mathbf{W} =\mathcal{E}_{1}(o) \cdot \mathbf{W}.
\end{equation}
Note that we omit the flattening operation that transforms a multi-dimensional matrix, $\mathcal{E}_{1}(g_1)$ or $\mathcal{E}_{1}(o)$, into a one-dimensional vector.
\end{tcolorbox}

%Let $\mathcal{F}$ denote the flattening operation that transforms a multi-dimensional matrix into a one-dimensional vector. 
\vspace*{-3mm}
\begin{proof}
The proof of Theorem 1 is based on the below lemmas. Please refer to Appendix (Section \ref{App: Lemma}) for the proofs of lemmas. We prove the Theorem 1 here.
\tcbset{colback=lightgray!15!white, colframe=white, left=1mm, right=1mm, top=1mm, bottom=1mm}
%\vspace*{-2mm}
\begin{tcolorbox}[breakable]
\textbf{Lemma 1.} \textit{Consider the diffusion model as defined in \textbf{Theorem 1}. Define $\bar{\alpha_t}$ as the key coefficient regulating the noise level. Let $\bm{\epsilon}$ denote the noise vector introduced during the diffusion process, and let  $\bm{\epsilon}_\theta(\mathbf{z}_t, t, \mathbf{c})$ represent the noise estimated by the diffusion model, where: $\theta$ denotes the parameters of the model, $\mathbf{z}_t$ represents the state of the system at time $t$, and $\mathbf{c}$ encapsulates the text-conditioning information. Under these conditions, the following identity holds:}
\begin{equation}
\mathcal{E}_{1}\left( g_{1} \right) - \mathcal{E}_{1}(o) =\frac{\sqrt{1-\bar{\alpha}_{t}}}{\sqrt{\bar{\alpha}_{t}}} \left( \bm{\epsilon} -\bm{\epsilon}_{\theta} \left( \mathbf{z}_{t} ,t,\mathbf{c} \right) \right).
\end{equation}
%\end{tcolorbox}
%\vspace*{-2mm}

%\tcbset{colback=lightgray!15!white, colframe=white, left=1mm, right=1mm, top=1mm, bottom=1mm}
%\vspace*{-2mm}
%\begin{tcolorbox}[breakable]
\textbf{Lemma 2.} \textit{Consider the equation $\mathbf{A X}=\mathbf{0}$, where $\mathbf{A}$ is a matrix. If $\mathbf{A}$ approximately equals to zero matrix, i.e., $\mathbf{A} \approx \mathbf{O}$, then there exists an approximate full-rank solution to the equation.}
\end{tcolorbox}
%\vspace*{-2mm}
%According to \textbf{Lemma 1}, we have:
%\begin{equation}
%\mathcal{E}_{1}\left( g_{1} \right) - \mathcal{E}_{1}(o) =\frac{\sqrt{1-\bar{\alpha}_{t}}}{\sqrt{\bar{\alpha}_{t}}} \left( \bm{\epsilon} -\bm{\epsilon}_{\theta} \left( \mathbf{z}_{t} ,t,\mathbf{c} \right) \right).
%\end{equation}
Because a well-trained diffusion model learns robust features and associations from diverse data, it generalizes well to inference prompts that are semantically similar to the training prompts. Moreover, the inference prompts here are generated by GPT-4o based on its understanding of the images, thus sharing semantic overlap with the training prompts. As a result, the estimated noise $\bm{\epsilon}_{\theta} \left( \mathbf{z}_{t} ,t,\mathbf{c} \right)$ closely approximates the true noise $\bm{\epsilon}$.
This means the difference between them is approximately equals to zero, \textit{i.e.,} $\bm{\epsilon} -\bm{\epsilon}_{\theta} \left( \mathbf{z}_{t} ,t,\mathbf{c} \right) \approx \mathbf{0}$. According to \textbf{Lemma 1}, this results in 
$\mathcal{E}_{1}\left( g_{1} \right) -\mathcal{E}_{1}(o)\approx \mathbf{0}$.
Denote $\mathbf{T}_1$ as the matrix, in which each column is $\mathcal{E}_{1}\left( g_{1} \right) -\mathcal{E}_{1}(o)$ from a training pair. According to \textbf{Lemma 2} and $\mathbf{T}_1 \approx \mathbf{O}$, we have $\mathbf{T}_1 \mathbf{X}=\mathbf{0}$ has an approximate full-rank solution. That means the matrix $\mathbf{W}$ satisfying Eq. \ref{Eq: T1} exists. 
\end{proof}
\vspace{-4mm}
\textbf{Note:} here we do \textbf{not} show that $\mathcal{E}_{1}(g_{1}) = \mathcal{E}_{1}(o)$ (in this case,  there would be no need of $\mathbf{W}$); instead, we prove that there exists a $\mathbf{W}$ that can further minimize the distance between $\mathcal{E}_{1}(g_{1})$ and $\mathcal{E}_{1}(o)$, despite the distance already being small. %Please see Table \ref{Table: performance} and Fig. \ref{Fig: cos} for experiments.

\subsection{Generalizability}
\tcbset{colback=lightroyalblue, colframe=white, left=1mm, right=1mm, top=1mm, bottom=1mm}
%\vspace*{-2mm}
\begin{tcolorbox}[breakable]
\textbf{Theorem 2.} \textit{Following \textbf{Theorem 1}, consider a different well-trained diffusion model $\mathcal{F}_2$ and its text-guided image-to-image functionability achieved by denoising noised images. The matrix $\mathbf{W}$ can be generalized such that for any original image $o$ and its translation $g_2$, we have:}
\begin{equation}\label{Eq: T2}
\mathcal{E}_{1}(g_2) \cdot \mathbf{W} = \mathcal{E}_{1}(o) \cdot \mathbf{W}.
\end{equation}
\end{tcolorbox}
\vspace*{-2mm}

\begin{proof} The proof of Theorem 2 is based on the below observation and lemmas. Please refer to Appendix (Section \ref{App: Lemma}) for the proofs of lemmas. We prove the Theorem 2 here.

\tcbset{colback=palegreen, colframe=white, left=1mm, right=1mm, top=1mm, bottom=1mm}
%\vspace*{-2mm}
\begin{tcolorbox}[breakable]
\textbf{Observation 1.} \textit{Consider two distinct matrices, $\mathbf{W}_1$ and $\mathbf{W}_2$, satisfying Eq. \ref{Eq: T1} and Eq. \ref{Eq: T2}, respectively. Let $\mathbf{v}_i$ denote the vector of all singular values of $\mathbf{W}_i$, where $i \in \{1, 2\}$. Specifically, define $\mathbf{v}_i = (\sigma_i^1, \sigma_i^2, \ldots, \sigma_i^k)$, with each $\sigma_i^j$ representing an singular value of $\mathbf{W}_i$. Despite the inequality $\mathbf{W}_1 \neq \mathbf{W}_2$, as shown in Table \ref{Table: cos}, it is observed that:}
\begin{equation}
cos\left( \varphi \right) =\frac{\mathbf{v}_{1} \cdot \mathbf{v}_{2}}{\| \mathbf{v}_{1} \| \| \mathbf{v}_{2} \|} \rightarrow 1.
\end{equation}
\end{tcolorbox}
%\vspace*{-2mm}

%\vspace{-2mm}

\tcbset{colback=lightgray!15!white, colframe=white, left=1mm, right=1mm, top=1mm, bottom=1mm}
%\vspace*{-2mm}
\begin{tcolorbox}
\textbf{Lemma 3 (Singular Value Decomposition).} \textit{Any matrix $\mathbf{A}$ can be decomposed into the product of three matrices: $\mathbf{A}=\mathbf{U} \mathbf{\Sigma} \mathbf{V^*}$, where $\mathbf{U}$ and $\mathbf{V}$ are orthogonal matrices, $\mathbf{\Sigma}$ is a diagonal matrix with non-negative singular values of $\mathbf{A}$  on the diagonal, and $\mathbf{V^*}$ is the conjugate transpose of $\mathbf{V}$.}
%\end{tcolorbox}
%\vspace*{-2mm}

%\tcbset{colback=lightgray!15!white, colframe=white, left=1mm, right=1mm, top=1mm, bottom=1mm}
%\begin{tcolorbox}
\textbf{Lemma 4.} \textit{A matrix $\mathbf{A}$ has a left inverse if and only if it has full rank.}
\end{tcolorbox}
%\vspace*{-2mm}
Consider $\mathbf{T_1}$ in the proof of \textbf{Theorem 1}, and denote $\mathbf{T_2}$ as the matrix, in which each column is $\mathcal{E}_1\left(g_2\right)-\mathcal{E}_1(o)$ from a training pair.
Therefore, we have $\mathbf{T}_{1} \mathbf{W}_{1} =\mathbf{0}$ and $\mathbf{T}_{2} \mathbf{W}_{2} =\mathbf{0}$. 
To prove \textbf{Theorem 2}, we only need to prove $\mathbf{T}_{2} \mathbf{W}_{1} =\mathbf{0}$.
According to \textbf{Lemma 3}, there exists orthogonal matrices, $\mathbf{U}_1$, $\mathbf{U}_2$, $\mathbf{V}_1$, and $\mathbf{V}_2$, with diagonal matrices,  $\mathbf{\Sigma}_1$ and $\mathbf{\Sigma}_2$, satisfying $\mathbf{W}_{1} =\mathbf{U}_{1} \Sigma_{1} \mathbf{V}_{1}^*$ and $\mathbf{W}_{2} =\mathbf{U}_{2} \Sigma_{2} \mathbf{V}_{2}^*$.
According to \textbf{Observation 1}, there exists $\alpha>0$ such that $\Sigma_{1} =\alpha \cdot \Sigma_{2}$.
Therefore, we have:
\begin{equation}
\begin{gathered}\mathbf{W}_{1} =\mathbf{U}_{1} \Sigma_{1} \mathbf{V}_{1}^{*} =\alpha \mathbf{U}_{1} \Sigma_{2} \mathbf{V}_{1}^{*}\\ =\alpha \mathbf{U}_{1} \left( \mathbf{U}_{2}^{\ast} \mathbf{W}_{2} \mathbf{V}_{2} \right) \mathbf{V}_{1}^{*} =\alpha \left( \mathbf{U}_{1} \mathbf{U}_{2}^{\ast} \right) \mathbf{W}_{2} \left( \mathbf{V}_{2} \mathbf{V}_{1}^{*} \right) .\end{gathered}
\end{equation}
Let $\mathbf{U}_{3} =\mathbf{U}_{1} \mathbf{U}_{2}^{\ast}$ and $\mathbf{V}_{3} =\mathbf{V}_{2} \mathbf{V}_{1}^{\ast}$, where $\mathbf{U}_{3}$ and $\mathbf{V}_{3}$ are thus orthogonal matrices. Therefore:
\begin{equation}
\begin{gathered}\parallel \mathbf{T}_{2} \mathbf{W}_{1} \parallel =\alpha \parallel \mathbf{T}_{2} \left( \mathbf{U}_{1} \mathbf{U}_{2}^{*} \right) \mathbf{W}_{2} \left( \mathbf{V}_{2} \mathbf{V}_{1}^{*} \right) \parallel\\ =\alpha \parallel \mathbf{T}_{2} \mathbf{U}_{3} \mathbf{W}_{2} \mathbf{V}_{3} \parallel\\ \leqslant \alpha \parallel \mathbf{T}_{2} \mathbf{U}_{3} \mathbf{W}_{2} \parallel \cdot \parallel \mathbf{V}_{3} \parallel =\alpha \parallel \mathbf{T}_{2} \mathbf{U}_{3} \mathbf{W}_{2} \parallel .\end{gathered}
\end{equation}
According to \textbf{Lemma 2 and 4}, there exists a matrix $\mathbf{K}$, such that $\mathbf{K} \mathbf{W}_2 = \mathbf{I}$. That means there exists $\mathbf{M}$, such that $\mathbf{U}_3 \mathbf{W}_2 = \mathbf{W}_2 \mathbf{M}$.
This results in: 
\begin{equation}
\begin{gathered}\left\| \mathbf{T}_{2} \mathbf{W}_{1} \right\| \leq \alpha \left\| \mathbf{T}_{2} \mathbf{U}_{3} \mathbf{W}_{2} \right\|\\ =\alpha \left\| \mathbf{T}_{2} \mathbf{W}_{2} \mathbf{M} \right\| \leq \alpha \left\| \mathbf{T}_{2} \mathbf{W}_{2} \right\| \cdot \| \mathbf{M} \|\end{gathered}
\end{equation}
Considering $\mathbf{T}_2 \mathbf{W}_2=\mathbf{0}$, we have $\mathbf{T}_2 \mathbf{W}_1=\mathbf{0}$.
\end{proof}

\begin{figure}[t]
    \centering
    \includegraphics[width=0.48\textwidth]{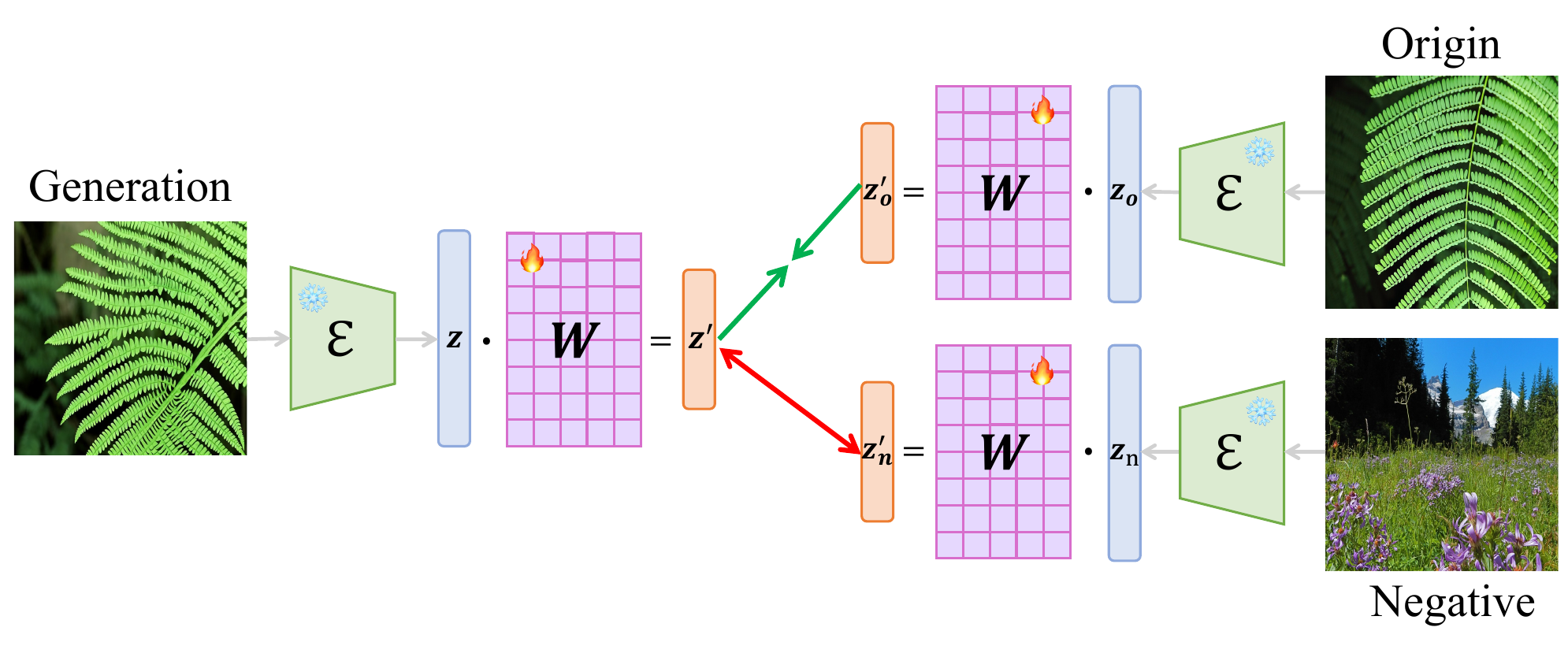}
    \vspace{-6mm}
    \caption{The implementation of learning theoretical-expected matrix $\mathbf{W}$. Specifically, in practice, we use gradient descent to optimize a metric loss in order to learn $\mathbf{W}$.} 
    \vspace{-2mm}
    \label{Fig: main}
\end{figure}

\vspace{-2mm}
\subsection{Implementation}

As illustrated in Fig. \ref{Fig: main}, we show how to learn the theoretical-expected matrix $\mathbf{W}$ in practice. Consider a triplet $\left( g, o, n \right)$, where $g$ is the generated image, $o$ is the origin used to generate $g$, and $n$ is a negative sample relative to $g$. We have:
\begin{equation}
\mathbf{z} =\mathcal{E} \left( g \right) ,\mathbf{z}_{o} =\mathcal{E} \left( o \right) ,\  \operatorname{and},\  \mathbf{z}_{n} =\mathcal{E} \left( n \right),
\end{equation}
where $\mathcal{E}$ is the encoder of VAE. Therefore, the final loss is defined as:
\begin{equation}\label{Eq: W}
\mathcal{L} =\mathcal{L}_{mtr} \left( \mathbf{z} \cdot \mathbf{W} ,\mathbf{z}_{o} \cdot \mathbf{W} ,\mathbf{z}_{n} \cdot \mathbf{W} \right),
\end{equation}
where $\mathcal{L}_{mtr}$ is a metric learning loss function that aims to bring positive data points closer together in the embedding space while pushing negative data points further apart. We use CosFace \citep{wang2018cosface} here as $\mathcal{L}_{mtr}$ for its simplicity and effectiveness. Using gradient descent, we can optimize the loss function $\mathcal{L}$ to obtain the theoretically expected matrix $\mathbf{W}$.

\section{Experiments}

\subsection{Protocols and Details}
\textbf{Evaluation protocols.} We adopt two commonly used evaluation metrics for our \tname~ task: \textit{i.e.}, Mean Average Precision (mAP) and Top-1 Accuracy (Acc). mAP evaluates a model’s precision at \textit{various} recall levels, while Acc measures the proportion of instances where the model’s \textit{top} prediction exactly matches the original image. Acc is stricter as it only counts when the first guess is correct.

\textbf{Dataset details.} During testing, the editing strengths for Stable Diffusion 2, Stable Diffusion XL, OpenDalle, ColorfulXL, Kandinsky-3, Stable Diffusion 3, and Kolors, are $0.9$, $0.8$, $0.7$, $0.7$, $0.6$, $0.8$, and $0.7$, respectively. The editing strengths used in testing are manually set to prevent significant visual differences between the generated images and the original ones. During training, the editing strength for Stable Diffusion 2 is $0.9$. The classifier-free guidance (CFG) scale is set to $7.5$ for all diffusion models, which is a commonly used value in practice.

\textbf{Training details.} We distribute the optimization of the theoretically expected matrix $\mathbf{W}$ across 8 NVIDIA A100 GPUs using PyTorch. The images are resized to a resolution of $256 \times 256$ before being embedded by the VAE encoder. The peak learning rate is set to $3.5 \times 10^{-4}$, and the Adam optimizer is used.

\begin{table}
        \caption{Publicly available models \textbf{fail} on the \dname~.}
\vspace*{-2mm}
\hspace{-2mm}
\small
\scalebox{1}{
  \begin{tabularx}{\hsize}{|>{\centering\arraybackslash}p{0.9cm}>{\raggedleft\arraybackslash}p{4.1cm}||Y|Y|}
    \hline\thickhline
    
   \rowcolor{mygray} &  \multirow{1}{*}{Method}& \scalebox{1}{mAP} & \scalebox{1}{Acc}  \\ \hline \hline
    \multirow{3 }{*}{Supervised}  & \scalebox{1.0}{Swin-B \citep{liu2021swin}} & $3.9$& $2.7$ \\
    \multirow{3 }{*}{Pre-trained}  & \scalebox{1.0}{ResNet-50 \citep{he2016deep}}  &$4.5$& $3.0$ \\ 
     \multirow{3 }{*}{Models}& \scalebox{1.0}{ConvNeXt \citep{liu2022convnet}} &$4.5$ & $3.1$ \\ 
     & \scalebox{1.0}{EfficientNet \citep{tan2019efficientnet}}   &$4.6$& $3.3$ \\
    & \scalebox{1.0}{ViT-B \citep{dosovitskiy2020vit}}  &$6.2$& $4.6$ \\

    \hline\hline
    \multirow{2 }{*}{Self-}& \scalebox{1.0}{SimSiam \citep{chen2021exploring}} &$1.8$ & $1.0$ \\
    \multirow{2 }{*}{supervised}& \scalebox{1.0}{MoCov3
     \citep{he2020momentum}} & $2.1$& $1.2$\\
     \multirow{2 }{*}{Learning}& \scalebox{1.0}{DINOv2 \citep{oquab2023dinov2}} & $4.3$& $2.9$ \\ 
     \multirow{2 }{*}{Models}& \scalebox{1.0}{MAE \citep{he2022masked}}  &$11.6$& $9.2$ \\
    & \scalebox{1.0}{SimCLR \citep{chen2020simple}} &$11.3$& $9.7$ \\

    \hline\hline
    \multirow{2}{*}{Vision-} & \scalebox{1.0}{CLIP \citep{radford2021learning}}  &$2.9$& $1.8$  \\ 
     \multirow{2}{*}{language}& \scalebox{1.0}{SLIP \citep{mu2022slip}}  &$5.4$& $3.7$ \\
    \multirow{2 }{*}{Models}& \scalebox{1.0}{ZeroVL \citep{cui2022contrastive}} &$5.6$& $3.8$ \\ 
    & \scalebox{1.0}{BLIP \citep{li2022blip}}  &$8.3$& $5.9$ 
     \\

\hline\hline
    
    \multirow{3 }{*}{Image Copy}& \scalebox{1.0}{ASL \citep{wang2023benchmark}}  &$5.2$& $4.1$ \\ 
    \multirow{3 }{*}{Detection}& \scalebox{1.0}{CNNCL \citep{yokoo2021contrastive}} &$6.3$& $5.0$ \\
     \multirow{3 }{*}{Models}& \scalebox{1.0}{BoT \citep{wang2021bag}}  &$10.5$& $8.2$ \\ 
    & \scalebox{1.0}{SSCD \citep{pizzi2022self}} & $14.8$& $12.5$ \\
    & \scalebox{1.0}{AnyPattern \citep{wang2024AnyPattern}}&$29.1$& $25.7$ \\
       \hline 
  \end{tabularx}}
  \label{Table: plausible}
   \vspace{-2mm}
\end{table}

\begin{figure}[t]
    \centering
    \includegraphics[width=0.48\textwidth]{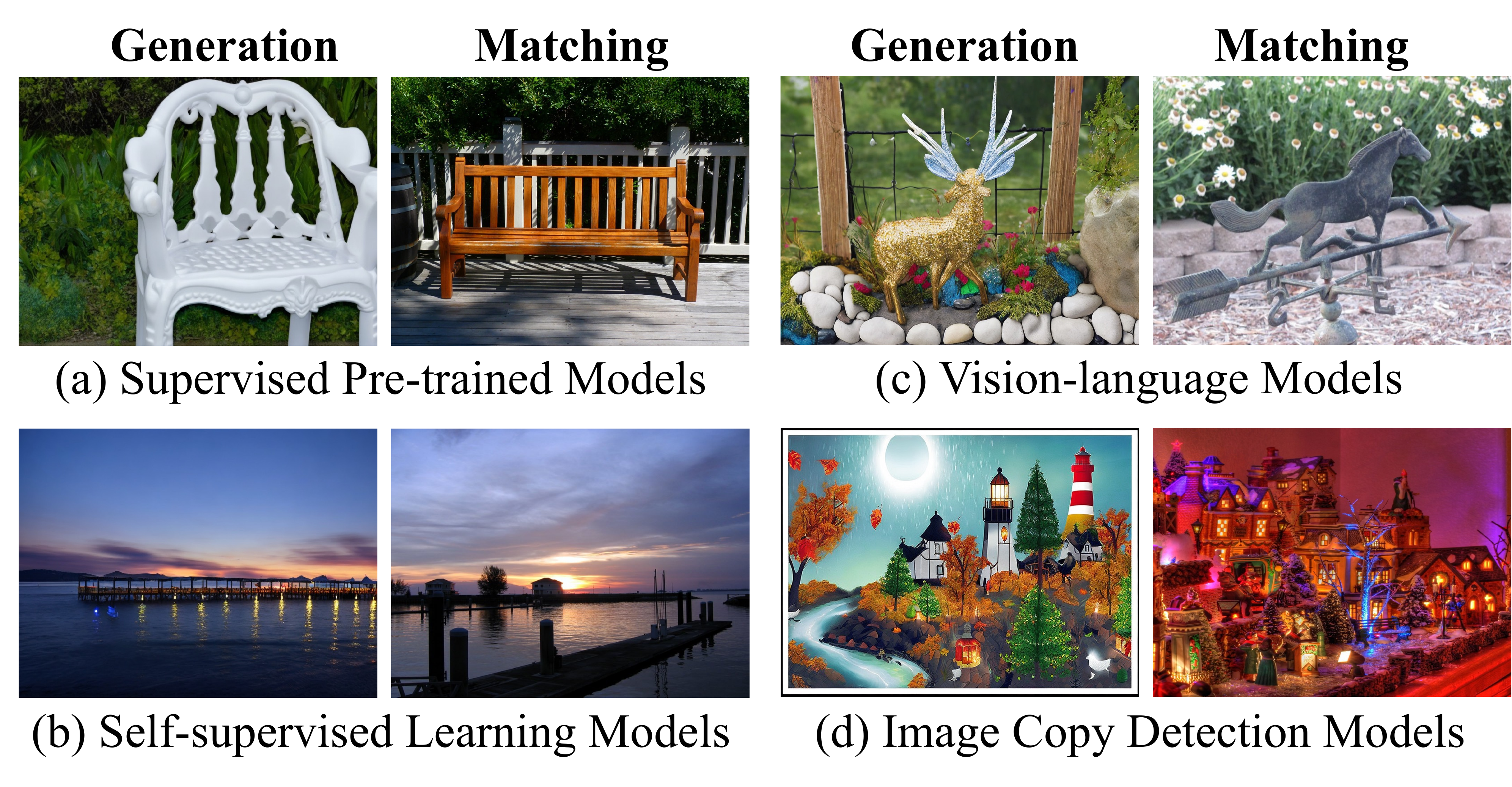}
    \vspace{-7mm}
    \caption{Examples of \textbf{failure} cases for each kind of model.} 
    \vspace{-5mm}
    \label{Fig: chall}
\end{figure}

\subsection{The Challenge from \tname~}
\begin{table*}[t]
\caption{Our method excels in \textbf{performance} while keeping \textbf{efficiency}. `mAP' and `Acc' are in percentage; `Train', `Extract', and `Match' are in `h', `$10^{-4}$ s/img', and `$10^{-10}$ s/pair', respectively.} 
\vspace*{-2mm}
\hspace*{-2mm}
\small
\scalebox{1}{
  \begin{tabularx}{\hsize}{|
>{\centering\arraybackslash}p{0.9cm}>{\raggedleft\arraybackslash}p{4.4cm}||Y|Y|Y|Y||Y|Y|Y||Y|Y|}
    \hline\thickhline
    
    \rowcolor{mygray}
 & &\multicolumn{2}{c|}{Seen $\uparrow$} 
     & \multicolumn{2}{c||}{Unseen $\uparrow$}
     & \multicolumn{3}{c||}{Efficiency $\downarrow$}&\multicolumn{2}{c|}{Manual edit $\uparrow$} \\ 
   \rowcolor{mygray} &  \multirow{-2}{*}{Method}& \scalebox{1.0}{mAP} & \scalebox{1.0}{Acc}  & \scalebox{1.0}{mAP}  & \scalebox{1.0}{Acc}&\scalebox{1.0}{Train}&\scalebox{1.0}{Extract}&\scalebox{1.0}{Match}& \scalebox{1.0}{mAP}  & \scalebox{1.0}{Acc}\\ \hline 
    \hline
    \multirow{-1 }{*}{Similarity}& \scalebox{1.0}{Circle loss \citep{sun2020circle}}&$70.4$& $64.3$ & $53.9$ & $48.5$&$1.79$&$2.81$&$0.80$&$76.6$&$74.5$\\ 
     \multirow{-1 }{*}{-based}& \scalebox{1.0}{SoftMax \citep{lecun1989backpropagation}}&$82.7$& $78.3$ & $55.0$ & $49.4$&$2.25$&$2.81$&$0.80$&$76.2$&$73.2$\\ 
    \multirow{-1 }{*}{Models}& \scalebox{1.0}{CosFace \citep{wang2018cosface}}& $87.1$& $83.2$ & $52.2$ & $46.5$ &$2.43$&$2.81$&$0.80$&$73.1$&$70.1$\\ 
%& \scalebox{1.0}{Upper: CosFace See All}&CVPR &\textcolor{black}{$90.0$}& \textcolor{black}{$86.8$} & \textcolor{gray}{$89.2$} & \textcolor{gray}{$86.2$}& $2.43$&$2.81$&$0.80$\\ 
    \hline\hline
     \multirow{-1 }{*}{General-}& \scalebox{1.0}{IBN-Net \citep{pan2018two}}&$88.6$& $85.1$ & $54.6$ & $49.0$&$2.03$&$3.42$&$2.14$&$75.4$&$72.1$\\ 
    \multirow{-1 }{*}{izable} & \scalebox{1.0}{TransMatcher \citep{liao2021transmatcher}}  &$65.6$& $60.3 $ & $65.3$ & $60.7$&$1.84$&$2.30$&$941$&$78.3$&$76.4$\\ 
   \multirow{-1 }{*}{Models}& \scalebox{1.0}{QAConv-GS \citep{liao2022graph}}& $78.8$ & $74.9$ & $75.8$ & $72.3$ &$1.47$&$2.30$&$464$&$74.4$&$71.9$\\ 
%& \scalebox{1.0}{Upper: QAConv-GS See All}&CVPR& \textcolor{black}{$78.8$}  & \textcolor{black}{$74.9$}   &\textcolor{gray}{$78.0$}   & \textcolor{gray}{$75.4$} &$1.47$&$2.30$&$464$\\ 
    \hline\hline
    & \scalebox{1.0}{Embeddings of VAE} &$51.0$& $47.0$ & $46.9$ & $43.0$&-&$1.59$&$4.25$&$66.6$&$64.6$\\ 
     \multirow{-1}{*}{\textbf{Ours}}& \cellcolor{lightroyalblue}\scalebox{1.0}{With Linear Transformation}&\cellcolor{lightroyalblue}$\mathbf{88.8}$& \cellcolor{lightroyalblue}$ \mathbf{86.6} $& \cellcolor{lightroyalblue}$ \mathbf{86.6} $& \cellcolor{lightroyalblue}$ \mathbf{84.5}$&\cellcolor{lightroyalblue}$ \mathbf{0.17}$&\cellcolor{lightroyalblue}$ \mathbf{1.59}$&\cellcolor{lightroyalblue}$ \mathbf{0.53}$&\cellcolor{lightroyalblue}$ \mathbf{86.6}$&\cellcolor{lightroyalblue}$ \mathbf{85.5}$\\ 
   & \scalebox{1.0}{Upper: Train\&Test Same Domain}& \textcolor{black}{$88.8$}& \textcolor{black}{$86.6$}& \textcolor{gray}{$92.0$} & \textcolor{gray}{$90.4$}&$0.17$&$1.59$&$0.53$&-&-\\ 
    \hline
  \end{tabularx}}
  \label{Table: performance}
  \vspace*{-2mm}
\end{table*}

\begin{figure*}[t]
    \centering
    \includegraphics[width=0.95\textwidth]{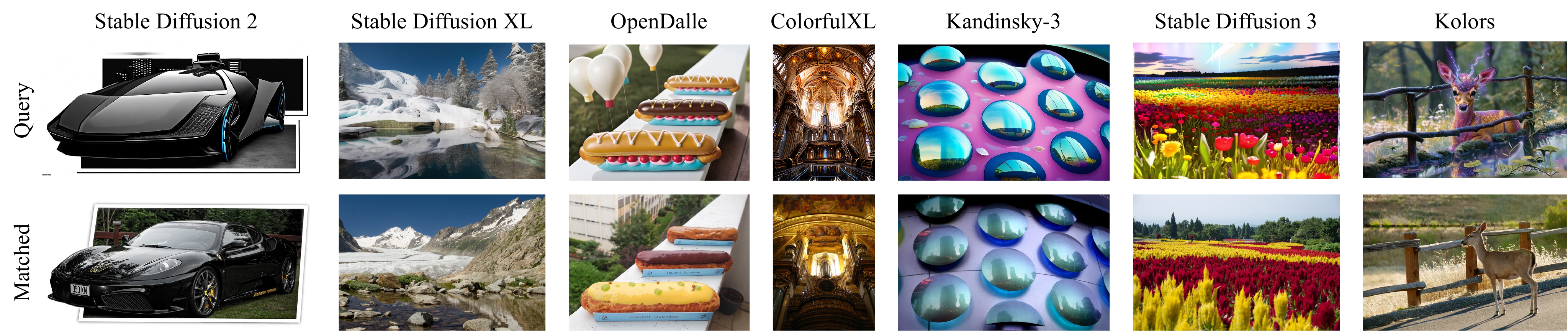}
    \vspace{-2mm}
    \caption{Examples of matchings achieved by our method. The origins can still be retrieved despite \textbf{non-trivial} alterations.} 
    \vspace{-5mm}
    \label{Fig: success}
\end{figure*}

\begin{figure*}[t]
    \centering
    \includegraphics[width=0.95\textwidth]{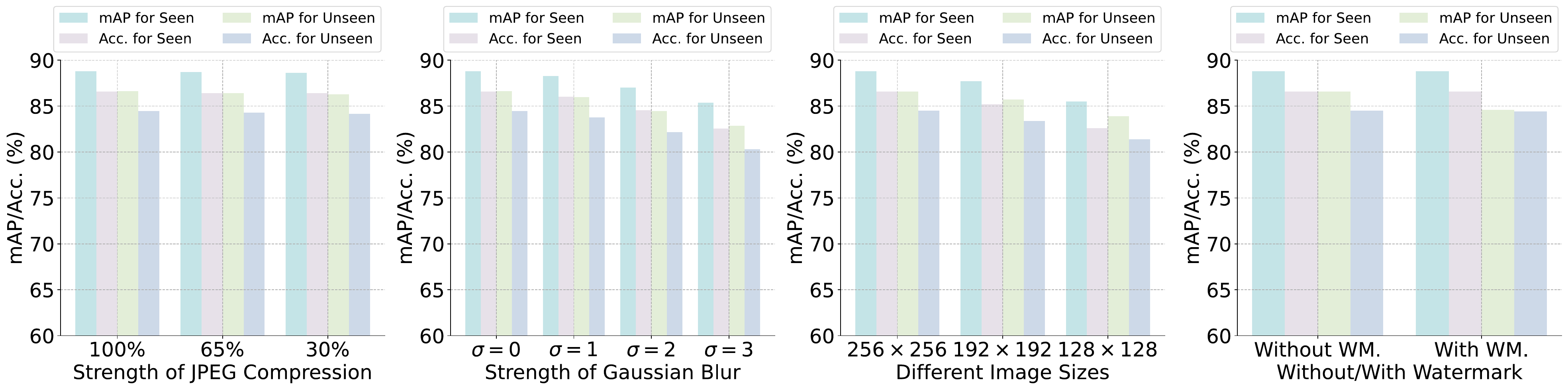}
    \vspace{-2mm}
    \caption{Our method demonstrates a certain level of \textbf{robustness} against different \textit{types} and \textit{intensities} of attacks.} 
    \vspace{-5mm}
    \label{Fig: robust}
\end{figure*}

\begin{table}[t]
\caption{VAE \textbf{differs} between seen and unseen models.} 
\vspace{-2mm}
%\hspace{-2mm}
\small
\scalebox{1}{
   \begin{tabularx}{\hsize}{|>{\centering\arraybackslash}p{0.1cm}>{\raggedleft\arraybackslash}p{0.95cm}||Y|Y|Y|Y|Y|Y|}
    \hline\thickhline
    
   \rowcolor{mygray}  & \multicolumn{1}{c||}{\hspace{3.5mm}Sim.}& \scalebox{1}{\hspace{-0.8mm}SDXL} & \scalebox{1}{\hspace{-0.8mm}OpDa}& \scalebox{1}{\hspace{-0.8mm}CoXL}& \scalebox{1}{\hspace{-0.4mm}Kan3}& \scalebox{1}{SD3}& \scalebox{1}{\hspace{-0.8mm}Kolor}  \\ \hline \hline
     &Conv.& $\hspace{-0.4mm}0.169$& $\hspace{-0.4mm}0.169$& $\hspace{-0.4mm}0.169$& $\hspace{-0.4mm}0.002$& -& $\hspace{-0.4mm}0.169$ \\
   \multirow{-2 }{*}{\vspace{-1mm}\rotatebox{90}{SD2}} &Embed.&$\hspace{-0.4mm}0.120$& $\hspace{-0.4mm}0.121$ & $\hspace{-0.4mm}0.120$& $\hspace{-0.4mm}0.023$& -& $\hspace{-0.4mm}0.120$\\ 
       \hline 
  \end{tabularx}}
  \label{Table: VAE}
\vspace{-6mm}
\end{table}
This section benchmarks popular public deep embedding models on the \dname~ test dataset. As shown in Table \ref{Table: plausible} and Fig. \ref{Fig: chall}, we extensively experiment on \textit{supervised pre-trained models}, \textit{self-supervised learning models}, \textit{vision-language models}, and \textit{image copy detection models}. We use these models as feature extractors, matching query features against references. The mAP and Acc are calculated by averaging the results of 7 diffusion models. Please refer to Table \ref{Table: all_1} in Appendix for the complete results. We observe that: \textbf{(1)} All existing methods \textbf{fail} on the \dname~ test dataset, highlighting the importance of constructing specialized training datasets and developing new methods. Specifically, \textit{supervised pre-trained models} overly focus on category-level similarity and thus achieve a maximum mAP of $6.2\%$; \textit{self-supervised learning models} handle only subtle changes and thus achieve a maximum mAP of $11.6\%$; \textit{vision-language models} return matches with overall semantic consistency, achieving a maximum mAP of $8.3\%$; and \textit{image copy detection models} are trained with translation patterns different from those of the \tname~ task, thus achieving a maximum mAP of $29.1\%$. \textbf{(2)} AnyPattern \citep{wang2024AnyPattern} achieves significantly higher mAP ($29.1\%$) and accuracy ($25.7\%$) compared to other methods. This is reasonable because it is designed for pattern generalization. Although the translation patterns generated by diffusion models in our \tname~ differ from the manually designed ones in AnyPattern, there remains some generalizability.

\subsection{VAE differs between Seen and Unseen Models} \label{Sec: VAE}
A common \textit{misunderstanding} is that the generalizability of our method comes from different diffusion models sharing the same or similar VAE. In Table \ref{Table: VAE}, we demonstrate that the VAE encoders used in our method \textbf{differ} between the diffusion models for generating training and testing images: \textbf{(1)} \textit{The parameters of VAE encoders are different.} For instance, the cosine similarity of the last convolutional layer weights of the VAE encoder between Stable Diffusion 2 and Stable Diffusion XL is only $0.169$. Furthermore, the number of channels in the last convolutional layer differs between Stable Diffusion 2 and Stable Diffusion 3. \textbf{(2)} \textit{The embeddings encoded by VAEs from different diffusion models vary.} For instance, the average cosine similarity of VAE embeddings for 100,000 original images between Stable Diffusion 2 and Kandinsky-3 is close to $0$. Additionally, the dimension of the VAE embedding for Stable Diffusion 2 is $4,096$, whereas for Stable Diffusion 3, it is $16,384$.

%\vspace{-6mm}
\subsection{The Effectiveness of our Method}
%\begin{figure}[t]
%    \centering
%    \includegraphics[width=0.471\textwidth]{cos.pdf}
%    \vspace{-4mm}
%    \caption{As expected by the theory, the cosine similarities \textbf{increase} w.r.t. epochs.} 
%    \vspace{-7mm}
%    \label{Fig: cos}
%\end{figure}

This section shows the effectiveness of our method in terms of (1) \textit{generalizability}, (2) \textit{efficiency}, (3) \textit{robustness}, and (4) the \textit{applicability} in the manual-editing scenarios. The experimental results for ‘Unseen’ are obtained by averaging the results from six different unseen diffusion models.

\textbf{Our method is much more generalizable than others.}
In Table \ref{Table: performance}, we compare our method with common similarity-based methods (incorporating domain generalization designs), all trained on the \dname~ training dataset. The mAP and Acc for `Unseen' are calculated by averaging the results of $6$ unseen diffusion models. Please refer to Table \ref{Table: all_2} for the complete results.
Fig. \ref{Fig: success} and Section \ref{App: fail} in the Appendix present the successful retrieval results and failure cases of our method, respectively.
We make three observations: \textbf{(1)} On \textbf{unseen} data, our method demonstrates significant performance superiority over common similarity-based models. Specifically, compared against the best one, we achieve a superiority of $+31.6\%$ mAP and $+35.1\%$ Acc. \textbf{(2)} Although domain generalization methods alleviate the generalization problem, they are still not satisfactory compared to ours (with at least a $-10.8\%$ mAP and $-9.1\%$ Acc). Moreover, those with the best performance suffer from severe efficiency issues, as detailed in the next section. \textbf{(3)} On the \textbf{seen} data, we achieve comparable performance with others. Specifically, there is a $0.2\%$ mAP and $1.5\%$ Acc superiority compared to the best one.

\textbf{Our method outperforms others in terms of efficiency.} Efficiency is crucial for the proposed task, as it often involves matching a query against a large-scale database in real-world scenarios. In Table \ref{Table: performance}, we compare the efficiency of our method with others regarding (1) \textit{training}, (2) \textit{feature extraction}, and (3) \textit{matching}. We draw three observations: \textbf{(1) Training:} Learning a matrix based on VAE embeddings is more efficient compared to training deep models on raw images. Specifically, our method is $8.6$ times faster than the nearest competitor. \textbf{(2) Feature extraction: } Compared to other models that use deep networks, such as ViT \citep{dosovitskiy2020vit}, the VAE encoder we use is relatively lightweight, resulting in faster feature extraction. \textbf{(3) Matching: } Compared to the best domain generalization models, QAConv-GS \citep{liao2022graph}, which use feature \textit{maps} for matching, our method still relies on feature \textit{vectors}. This leads to an $\mathbf{875} \boldsymbol{\times}$ superiority in matching speed.

\textbf{Our method is relatively robust against different attacks.} In the real world, the quality of an image may deteriorate during transmission. As shown in Fig. \ref{Fig: robust}, we apply varying intensities of JPEG compression, Gaussian blur, image resizing, and watermarking to evaluate the robustness of our method. It is observed that the side effects of these attacks are relatively minor. For instance, for the unseen diffusion models, the strongest Gaussian blur ($\sigma = 3$) reduces the mAP by only $3.7\%$, while the strongest compression ($30\%$) decreases the mAP by just $0.3\%$. Note that our models are \textbf{not} trained with these attacks.

\textbf{Our method is applicable to real-world scenarios with manually edited images.} 
In this section, we evaluate our method on a real-world dataset, SEED-Data-Edit \citep{ge2024seed}, which contains $52,000$ image editing samples. These samples were collected from amateur photographers who posted their images along with editing requests. Photoshop experts then fulfilled these requests, providing the edited images as target images.
Experimentally, we (1) de-duplicate to get $10,274$ image pairs; and (2) treat the edited (target) images as queries and search for them within a pool consisting of their origins along with $1,000,000$ distractor images.
The experiments in Table \ref{Table: performance} (right) show that: (1) our method generalizes effectively to real-world, manually edited images; and (2) it achieves the best performance compared to all competing methods.

%\textbf{Our training scheme successfully learns the theory-anticipated matrix $\mathbf{W}$.} In the theorems, we have proven that $\mathcal{E}_1\left(g_1/g_2\right) \cdot \mathbf{W} = \mathcal{E}_1(o) \cdot \mathbf{W}$ holds ideally. In Fig. \ref{Fig: cos}, we experimentally show this phenomenon. Specifically, we first calculate two cosine similarities of $<\mathcal{E}_{1} \left( g_{1} \right) \cdot \mathbf{W}, \mathcal{E}_{1}(o) \cdot \mathbf{W}>$ (seen) and $<\mathcal{E}_{1} \left( g_{2} \right) \cdot \mathbf{W}, \mathcal{E}_{1}(o) \cdot \mathbf{W}>$ (unseen), and then plot their changes with respect to the epochs. We observe that: \textbf{(1)} as expected, the two cosine similarities increase during training; and \textbf{(2)} the cosine similarities of the seen models are higher than those of the unseen ones, which is reasonable due to a certain degree of overfitting.

\subsection{Ablation Study}

%As shown in Table \ref{Table: Ablation-VAE}, Table \ref{Table: sup}, Fig. \ref{Fig: dims}, and Fig. \ref{Fig: layers},  we ablate the proposed method by (1) using different VAE \textit{encoders}, (2) supervising the training with different \textit{loss} functions, (3) exploring the minimum \textit{rank} of $\mathbf{W}$, and (4) experimentally exploring \textit{beyond} the theoretical guarantees. Please see the details in Appendix (Section \ref{Sup: Abla}).

In this section, we ablate the proposed method by (1) using different VAE \textit{encoders}, (2) supervising the training with different \textit{loss} functions, (3) exploring the minimum \textit{rank} of $\mathbf{W}$, (4) comparing against \textit{non-linear} transformations, and (5) analyzing the influence of editing \textit{strengths}.

%experimentally exploring \textit{beyond} the theoretical guarantees.

%As shown in Table \ref{Table: Ablation-VAE}, Table \ref{Table: sup}, Fig. \ref{Fig: dims}, and Fig. \ref{Fig: layers}, 
%We observe that: \textbf{(1)} Our method is insensitive to the choice of VAE encoder; \textbf{(2)} In practice, selecting an appropriate supervision for learning $\mathbf{W}$ is essential; \textbf{(3)} To improve efficiency, the rank of $\mathbf{W}$ can be relatively low; \textbf{(4)} Most non-linear transformations lead to overfitting; and \textbf{(5)} Our method maintains high performance across most editing strengths. Please see the details in Appendix (Section \ref{Sup: Abla}).

\begin{table*}[t]
\begin{minipage}{0.48\textwidth}
\caption{Ablation for choices of \textbf{VAE encoders}.} 
\vspace*{-2mm}
\hspace*{-2mm}
\small
\scalebox{1}{
  \begin{tabularx}{\hsize}{|>{\raggedleft\arraybackslash}p{2.3cm}||Y|Y|Y|Y|}
    \hline\thickhline
    \rowcolor{mygray}
 &\multicolumn{2}{c|}{Seen $\uparrow$} 
     & \multicolumn{2}{c|}{Unseen $\uparrow$} \\ 
\rowcolor{mygray}\multirow{-2}{*}{VAE}& \scalebox{1.0}{mAP} & \scalebox{1.0}{Acc}  & \scalebox{1.0}{mAP}  & \scalebox{1.0}{Acc}\\ \hline 
    
    \hline\hline
     \scalebox{1}{Open-Sora} &$86.3$& $83.5$ & $86.5$ & $84.2$\\ 
      \scalebox{1}{Open-Sora-Plan}&$88.8$& $86.4$ & $86.1$ & $84.0$\\ 
   \rowcolor{lightroyalblue} \scalebox{1}{Stable Diffusion 2} &$\mathbf{88.8}$& \cellcolor{lightroyalblue}$\mathbf{86.6}$ & \cellcolor{lightroyalblue}$\mathbf{86.6}$ & \cellcolor{lightroyalblue}$\mathbf{84.5}$\\ 
    \hline
  \end{tabularx}}
  \label{Table: Ablation-VAE}
  \vspace*{-2mm}
  \end{minipage}
  \hfill
\begin{minipage}{0.48\textwidth}
\caption{Ablation for different \textbf{supervision losses}.} 
\vspace*{-2mm}
\small
\scalebox{1}{
  \begin{tabularx}{\hsize}{|>{\raggedleft\arraybackslash}p{2.3cm}||Y|Y|Y|Y|}
    \hline\thickhline
    \rowcolor{mygray}
 &\multicolumn{2}{c|}{Seen $\uparrow$} 
     & \multicolumn{2}{c|}{Unseen $\uparrow$} \\ 
\rowcolor{mygray}\multirow{-2}{*}{Supervision}& \scalebox{1.0}{mAP} & \scalebox{1.0}{Acc}  & \scalebox{1.0}{mAP}  & \scalebox{1.0}{Acc}\\ \hline 
    
    \hline\hline
     \scalebox{1}{SoftMax} &$76.1$& $72.6$ & $62.4$ & $59.0$\\ 
      \scalebox{1}{Circle loss}&$84.9$& $82.0$ & $82.5$ & $80.4$\\ 
   \rowcolor{lightroyalblue} \scalebox{1}{CosFace}  &$\mathbf{88.8}$& \cellcolor{lightroyalblue}$\mathbf{86.6}$ & \cellcolor{lightroyalblue}$\mathbf{86.6}$ & \cellcolor{lightroyalblue}$\mathbf{84.5}$\\ 
    \hline
  \end{tabularx}}
  \label{Table: sup}
  \vspace*{-2mm}
  \end{minipage}

\end{table*}

\begin{figure*}[t]
\begin{minipage}{0.48\textwidth}
    \centering
    \includegraphics[width=0.95\textwidth]{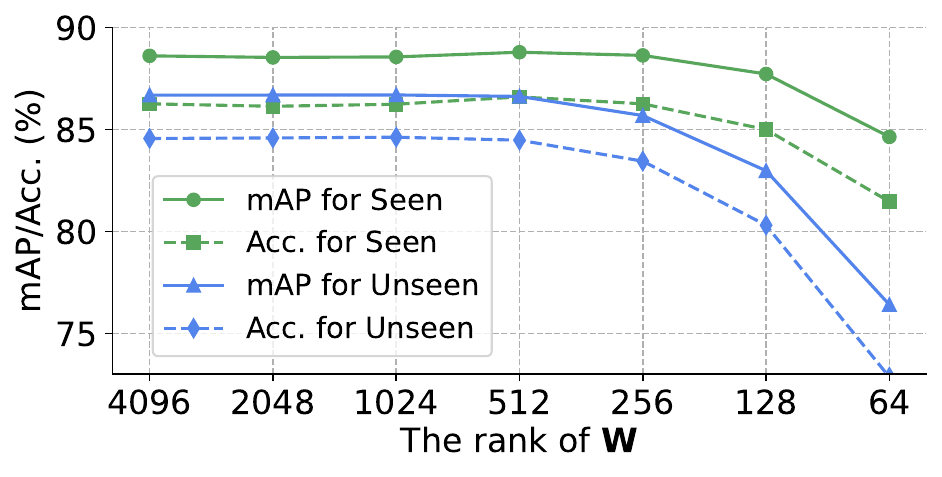}
    \vspace*{-4mm}
    \caption{The performance change w.r.t the rank of $\mathbf{W}$.} 
    \vspace*{-5mm}
    \label{Fig: dims}
 \end{minipage}
  \hfill
\begin{minipage}{0.48\textwidth}
    \centering
    \includegraphics[width=0.95\textwidth]{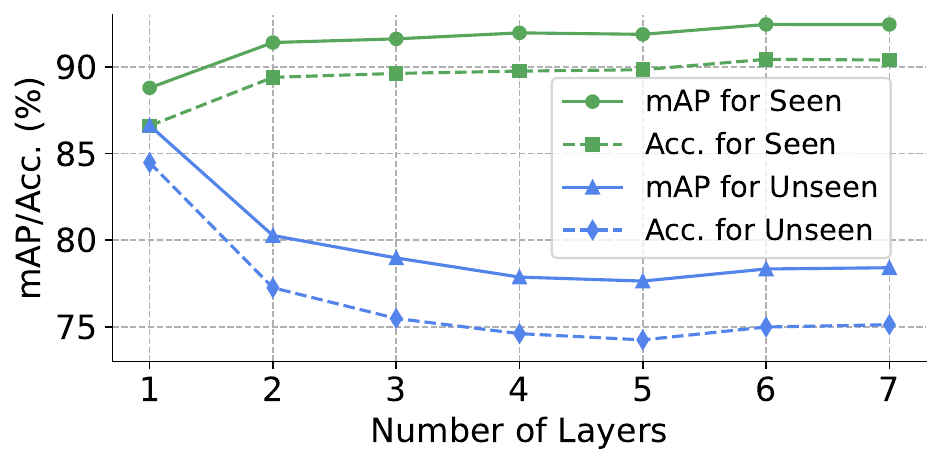}
    \vspace*{-4mm}
    \caption{The performance change w.r.t \textbf{number} of layers.} 
    \vspace*{-5mm}
    \label{Fig: layers}
    \end{minipage}
\end{figure*}

\begin{table}

\begin{minipage}{0.48\textwidth}
\caption{Comparison against \textbf{non-linear} transformations.} 
\vspace*{-2mm}
\small
\scalebox{1}{
  \begin{tabularx}{\hsize}{|>{\raggedleft\arraybackslash}p{2.3cm}||Y|Y|Y|Y|}
    \hline\thickhline
    \rowcolor{mygray}
 &\multicolumn{2}{c|}{Seen $\uparrow$} 
     & \multicolumn{2}{c|}{Unseen $\uparrow$} \\ 
\rowcolor{mygray}\multirow{-2}{*}{Transformation}& \scalebox{1.0}{mAP} & \scalebox{1.0}{Acc}  & \scalebox{1.0}{mAP}  & \scalebox{1.0}{Acc}\\ \hline 
    
    \hline\hline
      \scalebox{1}{Convolution}&$37.4$& $33.8$ & $32.5$ & $29.6$\\ 
      \scalebox{1}{Attention}&$89.0$& $87.2$ & $80.7$ & $78.2$\\ 
   \rowcolor{lightroyalblue} \scalebox{1}{Linear}  &$\mathbf{88.8}$& \cellcolor{lightroyalblue}$\mathbf{86.6}$ & \cellcolor{lightroyalblue}$\mathbf{86.6}$ & \cellcolor{lightroyalblue}$\mathbf{84.5}$\\ 
    \hline
  \end{tabularx}}
  \label{Table: nonlinear}
  \vspace*{-5mm}
  \end{minipage}

\end{table}
\textbf{Our method is insensitive to the choice of VAE encoder.} In Table \ref{Table: Ablation-VAE}, we replace the VAE encoder from Stable Diffusion 2 with two different encoders from Open-Sora \citep{opensora} and Open-Sora-Plan \citep{opensoraplan}. It is observed that, despite using significantly different well-trained VAEs, such as ones for videos, the performance drop is minimal (less than $1\%$).
This observation experimentally extends the Eq. \ref{Eq: T1} from $\mathcal{E}_1\left(g_1\right) \cdot \mathbf{W}=\mathcal{E}_1(o) \cdot \mathbf{W}$ to $\mathcal{E}_2\left(g_1\right) \cdot \mathbf{W}=\mathcal{E}_2(o) \cdot \mathbf{W}$, where $\mathcal{E}_{2}$ is an encoder from a totally different VAE.

\textbf{In practice, selecting an appropriate supervision for learning $\mathbf{W}$ is essential.} In Table \ref{Table: sup}, we replace the used supervision CosFace \citep{wang2018cosface} with two weaker supervisions, \textit{i.e.}, SoftMax \citep{lecun1989backpropagation} and Circle loss \citep{sun2020circle}. We observe that switching to Circle loss results in a drop in mAP for seen and unseen categories by $3.9\%$ and $4.1\%$, respectively. Furthermore, using SoftMax leads to mAP drops of $12.7\%$ and $24.2\%$ for the two categories, respectively. We infer this is because: while our theorems guarantee the distance between a translation and its origin, many negative samples serve as distractors during retrieval. Without appropriate hard negative solutions, these distractors compromise the final performance.

\textbf{To improve efficiency, the rank of $\mathbf{W}$ can be relatively low.} Assume the matrix $\mathbf{W}$ has a shape of $n \times m$, where $n$ is the dimension of the VAE embedding and $m$ is a hyperparameter. We show that $\mathbf{W}$ is approximately full-rank in the proof of \textit{existence}, and expect that $m \leq n$ in the proof of \textit{generalization}. Therefore, the rank of $\mathbf{W}$ is $m$. Experimentally, $n = 4,096$, and we explore the minimum rank of $\mathbf{W}$ from $4,096$ as shown in Fig. \ref{Fig: dims}. It is observed that: \textbf{(1)} From $4,096$ to $512$, the performance remains nearly unchanged. This suggests that we can train a relatively low-rank $\mathbf{W}$ to improve efficiency in real-world applications. \textbf{(2)} It is expected to see a performance decrease when reducing the rank from $512$ to $64$. This is because a matrix with too low rank cannot carry enough information to effectively linearly transform the VAE embeddings.

\textbf{Most non-linear transformations lead to overfitting.} In the theoretical section, we proved the existence and generalization of $\mathbf{W}$ using concepts from \textit{diffusion models} and \textit{linear algebra}. A natural experimental extension of this is to use an MLP with activation functions to replace the simple linear transformation ($\mathbf{W}$). Although linear algebra theory cannot guarantee these cases, we can still explore them experimentally. Experimentally, we increase the number of layers from $1$ to $7$, all using ReLU activation and residual connections. As shown in Fig. \ref{Fig: layers}, we observe overfitting in one type of diffusion model. Specifically, on one hand, the performance on seen diffusion models improves. For example, with 2 layers, the mAP increases to $91.4\%$ ($+2.6\%$), and Acc rises to $89.4\%$ ($+2.8\%$). However, on the other hand, a significant performance drop is observed on unseen diffusion models: with 2 layers, the mAP decreases from $86.6\%$ to $80.3\%$ ($-6.3\%$), and Acc drops from $84.5\%$ to $77.3\%$ ($-7.2\%$). The performance drop becomes even more severe when using more layers. Beyond that, we also try a single convolutional layer and a multi-head attention layer. The experiments in Table \ref{Table: nonlinear} show that: (1) likely due to underfitting, the simple convolutional layer results in a performance drop; and (2) although the multi-head attention layer marginally improves performance on seen images, its performance on unseen images falls behind our method, due to overfitting.

\textbf{Our method maintains high performance across most editing strengths.} Detailed experiments and analyses are provided in Appendix (Section \ref{App: Editing}).

%\subsection{Discussion}
%\vspace{-2mm}
%\begin{figure}[t]
 %   \centering
 %   \includegraphics[width=0.4\textwidth]{beyond.pdf}
  %  \vspace{-3mm}
  %  \caption{The image-to-image paradigm beyond our theorems.} 
 %   \vspace{-4mm}
   % \label{Fig: beyond}
%\end{figure}

\section{Conclusion}
This paper explores popular text-guided image-to-image diffusion models from a novel perspective: retrieving the original image of a query translated by these models. The proposed task, \tname~, is both important and timely, especially as awareness of security concerns posed by diffusion models grows. To support this task, we introduce the first \tname~ dataset, \dname~, designed with a focus on addressing generalization challenges. Specifically, the training set is generated by one diffusion model, while the test set is generated by seven different models. Furthermore, we propose a simple, generalizable solution with theoretical guarantees: First, we theoretically prove the existence of linear transformations that minimize the distance between the VAE embeddings of a query and its original image. Then, we demonstrate that the learned linear transformations generalize across different diffusion models, \textit{i.e.}, the VAE encoder and the learned transformations can effectively embed images generated by new diffusion models. %Finally, we hope this work helps combat misinformation, enforce copyright protection, and track target content in the real world.

%\textbf{Limitation.}
\vspace{-2mm}
\section*{Acknowledgment}
We sincerely thank OpenAI for their support through the
Researcher Access Program. Without their generous contribution, this work would not have been possible.

\section*{Impact Statement}
Our findings hold potentials for the responsible use of AI-generated content. Specifically, this research helps mitigate the growing concerns of misinformation, intellectual property violations, and content tracing evasion. 
However, the proposed model may produce false predictions. Therefore, the paper should not be interpreted as legal advice.

%\vspace{2mm}

\bibliography{main}

\begin{thebibliography}{49}
\providecommand{\natexlab}[1]{#1}
\providecommand{\url}[1]{\texttt{#1}}
\expandafter\ifx\csname urlstyle\endcsname\relax
  \providecommand{\doi}[1]{doi: #1}\else
  \providecommand{\doi}{doi: \begingroup \urlstyle{rm}\Url}\fi

\bibitem[Arkhipkin et~al.(2023)Arkhipkin, Filatov, Vasilev, Maltseva, Azizov,
  Pavlov, Agafonova, Kuznetsov, and Dimitrov]{arkhipkin2023kandinsky}
Arkhipkin, V., Filatov, A., Vasilev, V., Maltseva, A., Azizov, S., Pavlov, I.,
  Agafonova, J., Kuznetsov, A., and Dimitrov, D.
\newblock Kandinsky 3.0 technical report.
\newblock \emph{arXiv:2312.03511}, 2023.

\bibitem[Brooks et~al.(2023)Brooks, Holynski, and
  Efros]{brooks2023instructpix2pix}
Brooks, T., Holynski, A., and Efros, A.~A.
\newblock Instructpix2pix: Learning to follow image editing instructions.
\newblock In \emph{the IEEE/CVF Conference on Computer Vision and Pattern
  Recognition}, 2023.

\bibitem[Chen et~al.(2023)Chen, Wu, Lai, Ou, Liao, and
  Zheng]{chen2023challenges}
Chen, C., Wu, Z., Lai, Y., Ou, W., Liao, T., and Zheng, Z.
\newblock Challenges and remedies to privacy and security in aigc: Exploring
  the potential of privacy computing, blockchain, and beyond.
\newblock \emph{arXiv:2306.00419}, 2023.

\bibitem[Chen et~al.(2020)Chen, Kornblith, Norouzi, and Hinton]{chen2020simple}
Chen, T., Kornblith, S., Norouzi, M., and Hinton, G.
\newblock A simple framework for contrastive learning of visual
  representations.
\newblock In \emph{International conference on machine learning}, pp.\
  1597--1607. PMLR, 2020.

\bibitem[Chen \& He(2021)Chen and He]{chen2021exploring}
Chen, X. and He, K.
\newblock Exploring simple siamese representation learning.
\newblock In \emph{the IEEE/CVF conference on computer vision and pattern
  recognition}, 2021.

\bibitem[Cui et~al.(2022)Cui, Zhou, Guo, Yin, Wu, Yoshie, and
  Chen]{cui2022contrastive}
Cui, Q., Zhou, B., Guo, Y., Yin, W., Wu, H., Yoshie, O., and Chen, Y.
\newblock Contrastive vision-language pre-training with limited resources.
\newblock In \emph{European Conference on Computer Vision}, pp.\  236--253.
  Springer, 2022.

\bibitem[Dosovitskiy et~al.(2021)Dosovitskiy, Beyer, Kolesnikov, Weissenborn,
  Zhai, Unterthiner, Dehghani, Minderer, Heigold, Gelly, Uszkoreit, and
  Houlsby]{dosovitskiy2020vit}
Dosovitskiy, A., Beyer, L., Kolesnikov, A., Weissenborn, D., Zhai, X.,
  Unterthiner, T., Dehghani, M., Minderer, M., Heigold, G., Gelly, S.,
  Uszkoreit, J., and Houlsby, N.
\newblock An image is worth 16x16 words: Transformers for image recognition at
  scale.
\newblock \emph{ICLR}, 2021.

\bibitem[Esser et~al.(2024)Esser, Kulal, Blattmann, Entezari, M{\"u}ller,
  Saini, Levi, Lorenz, Sauer, Boesel, et~al.]{esser2024scaling}
Esser, P., Kulal, S., Blattmann, A., Entezari, R., M{\"u}ller, J., Saini, H.,
  Levi, Y., Lorenz, D., Sauer, A., Boesel, F., et~al.
\newblock Scaling rectified flow transformers for high-resolution image
  synthesis.
\newblock In \emph{Forty-first International Conference on Machine Learning},
  2024.

\bibitem[Fan et~al.(2023)Fan, Chen, Wang, and Huang]{fan2023trustworthiness}
Fan, M., Chen, C., Wang, C., and Huang, J.
\newblock On the trustworthiness landscape of state-of-the-art generative
  models: A comprehensive survey.
\newblock \emph{arXiv:2307.16680}, 2023.

\bibitem[Fernandez et~al.(2023)Fernandez, Douze, Jégou, and
  Furon]{fernandez2022active}
Fernandez, P., Douze, M., Jégou, H., and Furon, T.
\newblock Active image indexing.
\newblock In \emph{International Conference on Learning Representations
  (ICLR)}, 2023.

\bibitem[Ge et~al.(2024)Ge, Zhao, Li, Ge, and Shan]{ge2024seed}
Ge, Y., Zhao, S., Li, C., Ge, Y., and Shan, Y.
\newblock Seed-data-edit technical report: A hybrid dataset for instructional
  image editing.
\newblock \emph{arXiv preprint arXiv:2405.04007}, 2024.

\bibitem[He et~al.(2016)He, Zhang, Ren, and Sun]{he2016deep}
He, K., Zhang, X., Ren, S., and Sun, J.
\newblock Deep residual learning for image recognition.
\newblock In \emph{Proceedings of the IEEE conference on computer vision and
  pattern recognition}, pp.\  770--778, 2016.

\bibitem[He et~al.(2020)He, Fan, Wu, Xie, and Girshick]{he2020momentum}
He, K., Fan, H., Wu, Y., Xie, S., and Girshick, R.
\newblock Momentum contrast for unsupervised visual representation learning.
\newblock 2020.

\bibitem[He et~al.(2022)He, Chen, Xie, Li, Doll{\'a}r, and
  Girshick]{he2022masked}
He, K., Chen, X., Xie, S., Li, Y., Doll{\'a}r, P., and Girshick, R.
\newblock Masked autoencoders are scalable vision learners.
\newblock In \emph{Proceedings of the IEEE/CVF conference on computer vision
  and pattern recognition}, pp.\  16000--16009, 2022.

\bibitem[Izquierdo(2023)]{dataautogpt3}
Izquierdo, A.
\newblock Opendallev1.1, 2023.

\bibitem[KolorsTeam(2024)]{kolors}
KolorsTeam.
\newblock Kolors: Effective training of diffusion model for photorealistic
  text-to-image synthesis.
\newblock 2024.

\bibitem[LeCun et~al.(1989)LeCun, Boser, Denker, Henderson, Howard, Hubbard,
  and Jackel]{lecun1989backpropagation}
LeCun, Y., Boser, B., Denker, J.~S., Henderson, D., Howard, R.~E., Hubbard, W.,
  and Jackel, L.~D.
\newblock Backpropagation applied to handwritten zip code recognition.
\newblock \emph{Neural computation}, 1989.

\bibitem[Li et~al.(2022)Li, Li, Xiong, and Hoi]{li2022blip}
Li, J., Li, D., Xiong, C., and Hoi, S.
\newblock Blip: Bootstrapping language-image pre-training for unified
  vision-language understanding and generation.
\newblock In \emph{International Conference on Machine Learning}. PMLR, 2022.

\bibitem[Liao \& Shao(2021)Liao and Shao]{liao2021transmatcher}
Liao, S. and Shao, L.
\newblock Transmatcher: Deep image matching through transformers for
  generalizable person re-identification.
\newblock \emph{Advances in Neural Information Processing Systems},
  34:\penalty0 1992--2003, 2021.

\bibitem[Liao \& Shao(2022)Liao and Shao]{liao2022graph}
Liao, S. and Shao, L.
\newblock Graph sampling based deep metric learning for generalizable person
  re-identification.
\newblock In \emph{Proceedings of the IEEE/CVF Conference on Computer Vision
  and Pattern Recognition}, pp.\  7359--7368, 2022.

\bibitem[Lin et~al.(2024)Lin, Gupta, Zhang, Ren, Liu, Ding, Wang, Li,
  Verdoliva, and Hu]{lin2024detecting}
Lin, L., Gupta, N., Zhang, Y., Ren, H., Liu, C.-H., Ding, F., Wang, X., Li, X.,
  Verdoliva, L., and Hu, S.
\newblock Detecting multimedia generated by large ai models: A survey.
\newblock \emph{arXiv:2402.00045}, 2024.

\bibitem[Liu et~al.(2021)Liu, Lin, Cao, Hu, Wei, Zhang, Lin, and
  Guo]{liu2021swin}
Liu, Z., Lin, Y., Cao, Y., Hu, H., Wei, Y., Zhang, Z., Lin, S., and Guo, B.
\newblock Swin transformer: Hierarchical vision transformer using shifted
  windows.
\newblock \emph{the IEEE/CVF international conference on computer vision},
  2021.

\bibitem[Liu et~al.(2022)Liu, Mao, Wu, Feichtenhofer, Darrell, and
  Xie]{liu2022convnet}
Liu, Z., Mao, H., Wu, C.-Y., Feichtenhofer, C., Darrell, T., and Xie, S.
\newblock A convnet for the 2020s.
\newblock In \emph{Proceedings of the IEEE/CVF conference on computer vision
  and pattern recognition}, pp.\  11976--11986, 2022.

\bibitem[Meng et~al.(2022)Meng, He, Song, Song, Wu, Zhu, and Ermon]{mengsdedit}
Meng, C., He, Y., Song, Y., Song, J., Wu, J., Zhu, J.-Y., and Ermon, S.
\newblock Sdedit: Guided image synthesis and editing with stochastic
  differential equations.
\newblock In \emph{International Conference on Learning Representations}, 2022.

\bibitem[Mu et~al.(2022)Mu, Kirillov, Wagner, and Xie]{mu2022slip}
Mu, N., Kirillov, A., Wagner, D., and Xie, S.
\newblock Slip: Self-supervision meets language-image pre-training.
\newblock In \emph{European Conference on Computer Vision}, pp.\  529--544.
  Springer, 2022.

\bibitem[OpenAI(2024)]{openai2024gpt4o}
OpenAI.
\newblock Hello gpt-4o, 2024.

\bibitem[Oquab et~al.(2023)Oquab, Darcet, Moutakanni, Vo, Szafraniec, Khalidov,
  Fernandez, Haziza, Massa, El-Nouby, Howes, Huang, and
  et~al.]{oquab2023dinov2}
Oquab, M., Darcet, T., Moutakanni, T., Vo, H.~V., Szafraniec, M., Khalidov, V.,
  Fernandez, P., Haziza, D., Massa, F., El-Nouby, A., Howes, R., Huang, P.-Y.,
  and et~al.
\newblock Dinov2: Learning robust visual features without supervision, 2023.

\bibitem[Pan et~al.(2018)Pan, Luo, Shi, and Tang]{pan2018two}
Pan, X., Luo, P., Shi, J., and Tang, X.
\newblock Two at once: Enhancing learning and generalization capacities via
  ibn-net.
\newblock In \emph{Proceedings of the european conference on computer vision
  (ECCV)}, pp.\  464--479, 2018.

\bibitem[Papakipos et~al.(2022)Papakipos, Tolias, Jenicek, Pizzi, Yokoo, Wang,
  Sun, Zhang, Yang, Addicam, et~al.]{papakipos2022results}
Papakipos, Z., Tolias, G., Jenicek, T., Pizzi, E., Yokoo, S., Wang, W., Sun,
  Y., Zhang, W., Yang, Y., Addicam, S., et~al.
\newblock Results and findings of the 2021 image similarity challenge.
\newblock In \emph{NeurIPS 2021 Competitions and Demonstrations Track}, pp.\
  1--12. PMLR, 2022.

\bibitem[Pizzi et~al.(2022)Pizzi, Roy, Ravindra, Goyal, and
  Douze]{pizzi2022self}
Pizzi, E., Roy, S.~D., Ravindra, S.~N., Goyal, P., and Douze, M.
\newblock A self-supervised descriptor for image copy detection.
\newblock In \emph{the IEEE/CVF Conference on Computer Vision and Pattern
  Recognition}, 2022.

\bibitem[PKU-Yuan \& etc.(2024)PKU-Yuan and etc.]{opensoraplan}
PKU-Yuan, L. and etc., T.~A.
\newblock Open-sora-plan, April 2024.

\bibitem[Podell et~al.(2024)Podell, English, Lacey, Blattmann, Dockhorn,
  M{\"u}ller, Penna, and Rombach]{podell2024sdxl}
Podell, D., English, Z., Lacey, K., Blattmann, A., Dockhorn, T., M{\"u}ller,
  J., Penna, J., and Rombach, R.
\newblock {SDXL}: Improving latent diffusion models for high-resolution image
  synthesis.
\newblock In \emph{The Twelfth International Conference on Learning
  Representations}, 2024.

\bibitem[Radford et~al.(2021)Radford, Kim, Hallacy, Ramesh, Goh, Agarwal,
  Sastry, Askell, Mishkin, Clark, et~al.]{radford2021learning}
Radford, A., Kim, J.~W., Hallacy, C., Ramesh, A., Goh, G., Agarwal, S., Sastry,
  G., Askell, A., Mishkin, P., Clark, J., et~al.
\newblock Learning transferable visual models from natural language
  supervision.
\newblock In \emph{International conference on machine learning}, pp.\
  8748--8763. PMLR, 2021.

\bibitem[Recoilme(2023)]{ColorfulXL}
Recoilme.
\newblock Colorfulxl-lightning, 2023.

\bibitem[Rombach et~al.(2022)Rombach, Blattmann, Lorenz, Esser, and
  Ommer]{Rombach_2022_CVPR}
Rombach, R., Blattmann, A., Lorenz, D., Esser, P., and Ommer, B.
\newblock High-resolution image synthesis with latent diffusion models.
\newblock In \emph{Proceedings of the IEEE/CVF Conference on Computer Vision
  and Pattern Recognition (CVPR)}, pp.\  10684--10695, June 2022.

\bibitem[Sun et~al.(2020)Sun, Cheng, Zhang, Zhang, Zheng, Wang, and
  Wei]{sun2020circle}
Sun, Y., Cheng, C., Zhang, Y., Zhang, C., Zheng, L., Wang, Z., and Wei, Y.
\newblock Circle loss: A unified perspective of pair similarity optimization.
\newblock In \emph{Proceedings of the IEEE/CVF conference on computer vision
  and pattern recognition}, pp.\  6398--6407, 2020.

\bibitem[Tan \& Le(2019)Tan and Le]{tan2019efficientnet}
Tan, M. and Le, Q.
\newblock Efficientnet: Rethinking model scaling for convolutional neural
  networks.
\newblock In \emph{International conference on machine learning}, 2019.

\bibitem[Thomee et~al.(2016)Thomee, Shamma, Friedland, Elizalde, Ni, Poland,
  Borth, and Li]{thomee2016yfcc100m}
Thomee, B., Shamma, D.~A., Friedland, G., Elizalde, B., Ni, K., Poland, D.,
  Borth, D., and Li, L.-J.
\newblock Yfcc100m: The new data in multimedia research.
\newblock \emph{Communications of the ACM}, 59\penalty0 (2):\penalty0 64--73,
  2016.

\bibitem[Tumanyan et~al.(2023)Tumanyan, Geyer, Bagon, and
  Dekel]{tumanyan2023plug}
Tumanyan, N., Geyer, M., Bagon, S., and Dekel, T.
\newblock Plug-and-play diffusion features for text-driven image-to-image
  translation.
\newblock In \emph{the IEEE/CVF Conference on Computer Vision and Pattern
  Recognition}, 2023.

\bibitem[Wallace et~al.(2023)Wallace, Gokul, and Naik]{wallace2023edict}
Wallace, B., Gokul, A., and Naik, N.
\newblock Edict: Exact diffusion inversion via coupled transformations.
\newblock In \emph{Conference on Computer Vision and Pattern Recognition},
  2023.

\bibitem[Wang et~al.(2018)Wang, Wang, Zhou, Ji, Gong, Zhou, Li, and
  Liu]{wang2018cosface}
Wang, H., Wang, Y., Zhou, Z., Ji, X., Gong, D., Zhou, J., Li, Z., and Liu, W.
\newblock Cosface: Large margin cosine loss for deep face recognition.
\newblock In \emph{Proceedings of the IEEE conference on computer vision and
  pattern recognition}, pp.\  5265--5274, 2018.

\bibitem[Wang et~al.(2021)Wang, Zhang, Sun, and Yang]{wang2021bag}
Wang, W., Zhang, W., Sun, Y., and Yang, Y.
\newblock Bag of tricks and a strong baseline for image copy detection.
\newblock \emph{arXiv:2111.08004}, 2021.

\bibitem[Wang et~al.(2023)Wang, Sun, and Yang]{wang2023benchmark}
Wang, W., Sun, Y., and Yang, Y.
\newblock A benchmark and asymmetrical-similarity learning for practical image
  copy detection.
\newblock In \emph{Proceedings of the AAAI Conference on Artificial
  Intelligence}, volume~37, pp.\  2672--2679, 2023.

\bibitem[Wang et~al.(2024{\natexlab{a}})Wang, Sun, Tan, and
  Yang]{wang2024AnyPattern}
Wang, W., Sun, Y., Tan, Z., and Yang, Y.
\newblock Anypattern: Towards in-context image copy detection.
\newblock In \emph{arXiv:2404.13788}, 2024{\natexlab{a}}.

\bibitem[Wang et~al.(2024{\natexlab{b}})Wang, Sun, and Yang]{wang2024pattern}
Wang, W., Sun, Y., and Yang, Y.
\newblock Pattern-expandable image copy detection.
\newblock \emph{International Journal of Computer Vision}, pp.\  1--17,
  2024{\natexlab{b}}.

\bibitem[Wang et~al.(2024{\natexlab{c}})Wang, Sun, Yang, Hu, Tan, and
  Yang]{wang2024replication}
Wang, W., Sun, Y., Yang, Z., Hu, Z., Tan, Z., and Yang, Y.
\newblock Replication in visual diffusion models: A survey and outlook.
\newblock \emph{arXiv:2408.00001}, 2024{\natexlab{c}}.

\bibitem[Ye et~al.(2023)Ye, Zhang, Liu, Han, and Yang]{ye2023ip-adapter}
Ye, H., Zhang, J., Liu, S., Han, X., and Yang, W.
\newblock Ip-adapter: Text compatible image prompt adapter for text-to-image
  diffusion models.
\newblock 2023.

\bibitem[Yokoo(2021)]{yokoo2021contrastive}
Yokoo, S.
\newblock Contrastive learning with large memory bank and negative embedding
  subtraction for accurate copy detection.
\newblock \emph{arXiv:2112.04323}, 2021.

\bibitem[Zheng et~al.(2024)Zheng, Peng, Yang, Shen, Li, Liu, Zhou, Li, and
  You]{opensora}
Zheng, Z., Peng, X., Yang, T., Shen, C., Li, S., Liu, H., Zhou, Y., Li, T., and
  You, Y.
\newblock Open-sora: Democratizing efficient video production for all, March
  2024.

\end{thebibliography}
\bibliographystyle{icml2025}

%%%%%%%%%%%%%%%%%%%%%%%%%%%%%%%%%%%%%%%%%%%%%%%%%%%%%%%%%%%%%%%%%%%%%%%%%%%%%%%
%%%%%%%%%%%%%%%%%%%%%%%%%%%%%%%%%%%%%%%%%%%%%%%%%%%%%%%%%%%%%%%%%%%%%%%%%%%%%%%
% APPENDIX
%%%%%%%%%%%%%%%%%%%%%%%%%%%%%%%%%%%%%%%%%%%%%%%%%%%%%%%%%%%%%%%%%%%%%%%%%%%%%%%
%%%%%%%%%%%%%%%%%%%%%%%%%%%%%%%%%%%%%%%%%%%%%%%%%%%%%%%%%%%%%%%%%%%%%%%%%%%%%%%
\newpage
\appendix
\begin{figure*}[t]
    \centering
    \includegraphics[width=0.94\textwidth]{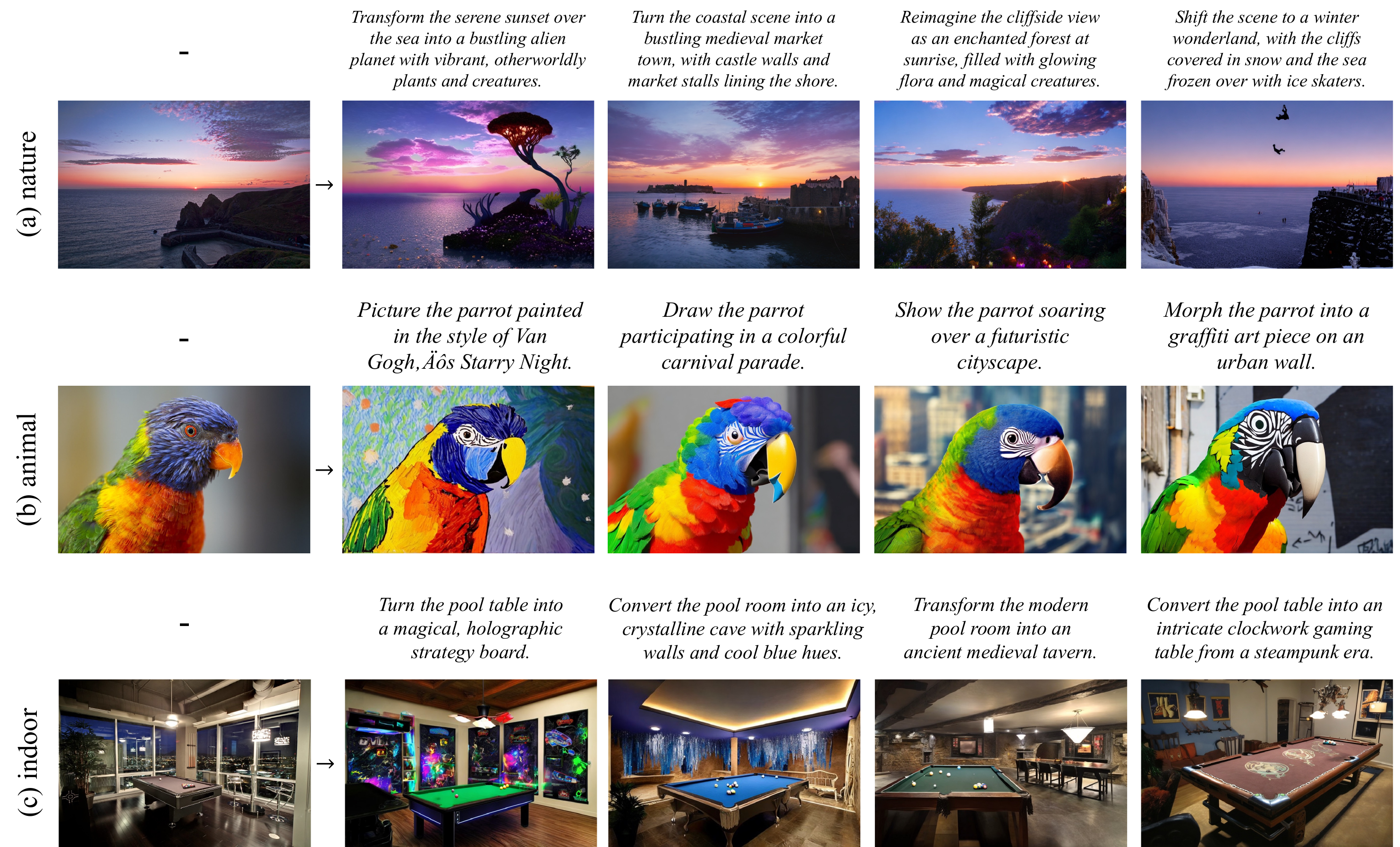}
    %\vspace{-1mm}
    \caption{Illustration of prompts and corresponding generated images for 3 different subjects in our dataset. Our dataset comprehensively includes various subjects found in the real world.} 
     \vspace{-4mm}
    \label{Fig: prompt}
\end{figure*}
\section{Proofs of Lemmas}\label{App: Lemma}
\tcbset{colback=lightgray!15!white, colframe=white, left=1mm, right=1mm, top=1mm, bottom=1mm}
%\vspace*{-2mm}
\begin{tcolorbox}
\textbf{Lemma 1.} \textit{Consider the diffusion model as defined in \textbf{Theorem 1}. Define $\bar{\alpha_t}$ as the key coefficient regulating the noise level. Let $\bm{\epsilon}$ denote the noise vector introduced during the diffusion process, and let  $\bm{\epsilon}_\theta(\mathbf{z}_t, t, \mathbf{c})$ represent the noise estimated by the diffusion model, where: $\theta$ denotes the parameters of the model, $\mathbf{z}_t$ represents the state of the system at time $t$, and $\mathbf{c}$ encapsulates the text-conditioning information. Under these conditions, the following identity holds:}
\begin{equation}\label{App: eq1}
\mathcal{E}_{1}\left( g_{1} \right) - \mathcal{E}_{1}(o) =\frac{\sqrt{1-\bar{\alpha}_{t}}}{\sqrt{\bar{\alpha}_{t}}} \left( \bm{\epsilon} -\bm{\epsilon}_{\theta} \left( \mathbf{z}_{t} ,t,\mathbf{c} \right) \right).
\end{equation}
\end{tcolorbox}
\vspace*{-2mm}

\begin{proof}
Denote $\mathbf{z}_{0} =\mathcal{E}_{1}(o)$ and $\mathbf{z}_{0}^{\prime}$ as $\mathbf{z}_{0}$ after adding noise and denoising. Therefore, we have
\begin{equation}
\mathcal{E}_{1}\left( g_{1} \right) -\mathcal{E}_{1}(o)=\mathcal{E}_{1}\left( \mathcal{D}_{1}\left( \mathbf{z}_{0}^{\prime} \right) \right) -\mathbf{z}_{0} =\mathbf{z}_{0}^{\prime} -\mathbf{z}_{0},
\end{equation}
where $\mathcal{D}_{1}$ is the decoder of VAE.

Given an initial data point $\mathbf{z}_0$, the forward process in a diffusion model adds noise to the data step by step. The expression for $\mathbf{z}_t$ at a specific timestep $t$ can be written as:
\begin{equation}
\mathbf{z}_t=\sqrt{\bar{\alpha_t}} \mathbf{z}_0+\sqrt{1-\bar{\alpha_t}} \bm{\epsilon}.
\end{equation}
To denoise $\mathbf{z}_t$ and recover an estimate of the original data $\mathbf{z}_0$, the reverse process is used. A neural network $\theta$ is trained to predict the noise $\bm{\epsilon}$ added to $\mathbf{z}_0$. The denoised data $\mathbf{z}_0^{\prime}$ can be expressed as:
\begin{equation}
\mathbf{z}_0^{\prime}=\frac{1}{\sqrt{\bar{\alpha}_t}}\left(\mathbf{z}_t-\sqrt{1-\bar{\alpha}_t} \bm{\epsilon}_\theta\left(\mathbf{z}_t, t,\mathbf{c}\right)\right).
\end{equation}
Therefore, we have:
\begin{equation}
\begin{gathered}\mathcal{E}_{1}\left( g_{1} \right) -\mathcal{E}_{1}(o)=\mathbf{z}_{0}^{\prime} -\mathbf{z}_{0}\\ =\frac{1}{\sqrt{\bar{\alpha}_{t}}} \left( \mathbf{z}_{t} -\sqrt{1-\bar{\alpha}_{t}} \bm{\epsilon}_{\theta} \left( \mathbf{z}_{t} ,t ,\mathbf{c}\right) \right) -\mathbf{z}_{0}\\ =\frac{1}{\sqrt{\bar{\alpha}_{t}}} \left( \sqrt{\bar{\alpha}_{t}} \mathbf{z}_{0} +\sqrt{1-\bar{\alpha}_{t}} \bm{\epsilon} -\sqrt{1-\overline{\alpha_{t}}} \bm{\epsilon}_{\theta} \left( \mathbf{z}_{t} ,t,\mathbf{c} \right) \right) -\mathbf{z}_{0}\\ =\frac{\sqrt{1-\bar{\alpha}_{t}}}{\sqrt{\bar{\alpha}_{t}}} \left( \bm{\epsilon} -\bm{\epsilon}_{\theta} \left( \mathbf{z}_{t} ,t,\mathbf{c} \right) \right) .\end{gathered}
\end{equation}
The Eq. \ref{App: eq1} is proved.
\end{proof}

\tcbset{colback=lightgray!15!white, colframe=white, left=1mm, right=1mm, top=1mm, bottom=1mm}
\vspace*{-2mm}
\begin{tcolorbox}
\textbf{Lemma 2.} \textit{Consider the equation $\mathbf{A X}=\mathbf{0}$, where $\mathbf{A}$ is a matrix. If $\mathbf{A}$ approximately equals to zero matrix, i.e., $\mathbf{A} \approx \mathbf{O}$, then there exists an approximate full-rank solution to the equation.}
\end{tcolorbox}
\vspace*{-2mm}
\begin{proof}
Consider a matrix $\mathbf{A} \in \mathbb{R}^{m\times n}$.
According to \textbf{Lemma 3}, there exists orthogonal matrices $\mathbf{U}\in \mathbb{R}^{m\times m} $ and $\mathbf{V}\in \mathbb{R}^{n\times n} $, and diagonal matrix $\mathbf{\Sigma} \in \mathbb{R}^{m\times n}$ with non-negative singular values, such that, $\mathbf{A}=\mathbf{U} \mathbf{\Sigma} \mathbf{V}^*$. Therefore, the linear equation can be transformed as:
\begin{equation}
\mathbf{U} \mathbf{\Sigma} \mathbf{V}^*\mathbf{X} = \mathbf{0}.
\end{equation}
Considering $\mathbf{U}^* \mathbf{U} = \mathbf{I}$ and denoting $\mathbf{X}^{\prime} = \mathbf{V}^* \mathbf{X}$, we have 
$ \mathbf{\Sigma} \mathbf{X}^{\prime} = \mathbf{0}$. Because $\mathbf{A} \approx \mathbf{O}$, all of its singular values approximately equals to $0$. Considering the floating-point precision we need, $ \mathbf{\Sigma} \mathbf{X}^{\prime} = \mathbf{0}$ could be regarded as:
\begin{equation}
\left( \begin{gathered}\left( \begin{matrix}\sigma_{0}&&&\\ &\sigma_{1}&&\\ &&\ddots&\\ &&&\sigma_{r}\end{matrix} \right) \left( \begin{matrix}0&&&\\ &0&&\\ &&\ddots&\\ &&&0\end{matrix} \right)\\ \left( \begin{matrix}0&&&\\ &0&&\\ &&\ddots&\\ &&&0\end{matrix} \right) \  \left( \begin{matrix}0&&&\\ &0&&\\ &&\ddots&\\ &&&0\end{matrix} \right)\end{gathered} \right) \mathbf{X} ^\prime =\mathbf{0},
\end{equation}
where $r$ is the number of non-zero singular values.
Therefore, there exists an $\mathbf{Z}^{\prime} \in \mathbb{R}^{n\times k}$ with $\text{rank} = \operatorname{min}(m,n) - r$. When $k\leqslant \operatorname{min}(m,n) - r$, $\mathbf{Z}^{\prime}$ is full rank, \textit{i.e}, $\mathbf{Z} = \mathbf{V} \mathbf{Z}^{\prime}$ is an approximate full-rank solution to the linear equation $\mathbf{A X}=\mathbf{0}$.
\end{proof}

\tcbset{colback=lightgray!15!white, colframe=white, left=1mm, right=1mm, top=1mm, bottom=1mm}
\vspace*{-2mm}
\begin{tcolorbox}
\textbf{Lemma 3 (Singular Value Decomposition).} \textit{Any matrix $\mathbf{A}$ can be decomposed into the product of three matrices: $\mathbf{A}=\mathbf{U} \mathbf{\Sigma} \mathbf{V^*}$, where $\mathbf{U}$ and $\mathbf{V}$ are orthogonal matrices, $\mathbf{\Sigma}$ is a diagonal matrix with non-negative singular values of $\mathbf{A}$  on the diagonal, and $\mathbf{V^*}$ is the conjugate transpose of $\mathbf{V}$.}
\end{tcolorbox}
\vspace*{-2mm}

\begin{proof}
Consider a matrix $\mathbf{A} \in \mathbb{R}^{m\times n}$. The matrix $\mathbf{A}^{*} \mathbf{A}$ is therefore symmetric and positive semi-definite, which means the matrix is diagonalizable with an eigendecomposition of the form:
\begin{equation}
\mathbf{A}^{*} \mathbf{A}=\mathbf{V} \Lambda \mathbf{V}^{*}=\sum_{i=1}^n \lambda_i \mathbf{v}_i \mathbf{v}_i^{*}=\sum_{i=1}^n\left(\sigma_i\right)^2 \mathbf{v}_i \mathbf{v}_i^{*},
\end{equation}
where $\mathbf{V}$ is an orthonormal matrix whose columns are the eigenvectors of  $\mathbf{A}^{*} \mathbf{A}$. 

We have defined the singular value $\sigma_i$ as the square root of the $i$-th eigenvalue; we know we can take the square root of our eigenvalues because positive semi-definite matrices can be equivalently characterized as matrices with non-negative eigenvalues.

For the $i$-th eigenvector-eigenvalue pair, we have
\begin{equation}
\mathbf{A}^{*} \mathbf{A} \mathbf{v}_i=\left(\sigma_i\right)^2 \mathbf{v}_i.
\end{equation}
Define a new vector $\mathbf{u}_i$, such that,
\begin{equation}
\mathbf{u}_i=\frac{\mathbf{A}\mathbf{v}_i}{\sigma_i}.
\end{equation}
This construction enables $\mathbf{u}_i$ as a unit eigenvector of $\mathbf{A} \mathbf{A}^{*}$.
Now let $\mathbf{V}$ be an $n \times n$ matrix -- because $\mathbf{A} \mathbf{A}^{*}$ is $n \times n$ -- where the $i$-th column is $\mathbf{v}_i$; let $\mathbf{U}$ be an $m \times m$ matrix -- because $\mathbf{A} \mathbf{v}_i$ is an $m$-vector -- where the $i$-th column is $\mathbf{u}_i$; and let $\mathbf{\Sigma}$ be a diagonal matrix whose $i$-th element is $\sigma_i$. Then we can express the relationships we have so far in matrix form as:
\begin{equation}
\begin{aligned} \mathbf{U} & =\mathbf{A} \mathbf{V} \mathbf{\Sigma}^{-1}, \\ \mathbf{U} \mathbf{\Sigma} & =\mathbf{A} \mathbf{V}, \\ \mathbf{A} & =\mathbf{U} \mathbf{\Sigma} \mathbf{V}^{*},\end{aligned}
\end{equation}
where we use the fact that $\mathbf{V} \mathbf{V}^{*}=I$ and $\mathbf{\Sigma}^{-1}$ is a diagonal matrix where the $i$-th value is the reciprocal of $\sigma_i$. 
\end{proof}

\tcbset{colback=lightgray!15!white, colframe=white, left=1mm, right=1mm, top=1mm, bottom=1mm}
\vspace*{-2mm}
\begin{tcolorbox}
\textbf{Lemma 4.} \textit{A matrix $\mathbf{A}$ has a left inverse if and only if it has full rank.}
\end{tcolorbox}
\vspace*{-2mm}

\begin{proof}
To prove Lemma 4, we must demonstrate two directions: if a matrix $\mathbf{A}$ has a left inverse, then it must have full rank, and conversely, if a matrix $\mathbf{A}$ has full rank, then it has a left inverse.

(1) Suppose $\mathbf{A} \in \mathbb{R}^{m\times n}$ has a left inverse $\mathbf{B} \in \mathbb{R}^{n\times m}$ such that $\mathbf{B} \mathbf{A} = \mathbf{I}_n$. Because the $ \mathbf{I}_n$ is of rank $n$, the matrix $\mathbf{A}\mathbf{B}$ must have rank $n$. Considering the inequality: 
\begin{equation}
\begin{gathered}n=\text{rank} (\mathbf{BA} )\leq \min (\text{rank} (\mathbf{A} ),\text{rank} (\mathbf{B} ))\\ \leq \text{rank} (\mathbf{A} )\leq \min (m,n)\leq n\end{gathered}
\end{equation}
we have $\text{rank}(\mathbf{A})=n$, \textit{i.e.}, $\mathbf{A}$ has full rank. 

(2) Suppose $\mathbf{A} \in \mathbb{R}^{m\times n}$ has full rank, \textit{i.e.}, $\text{rank}(\mathbf{A}) = \min(m, n)$. We have the rows of $\mathbf{A}$ are linearly independent, and thus there exists an $n \times m$ matrix $\mathbf{C}$ such that $\mathbf{C} \mathbf{A} = \mathbf{I}_n $. That means $\mathbf{C}$ is a left inverse of $\mathbf{A}$.
\end{proof}

\section{Implementation of GPT-4o}\label{App: gpt}
\begin{figure}[t]

    %\centering
    \hspace{-3mm}
    \begin{minipage}{0.19\textwidth}
        \centering
        \includegraphics[height=1.5cm]{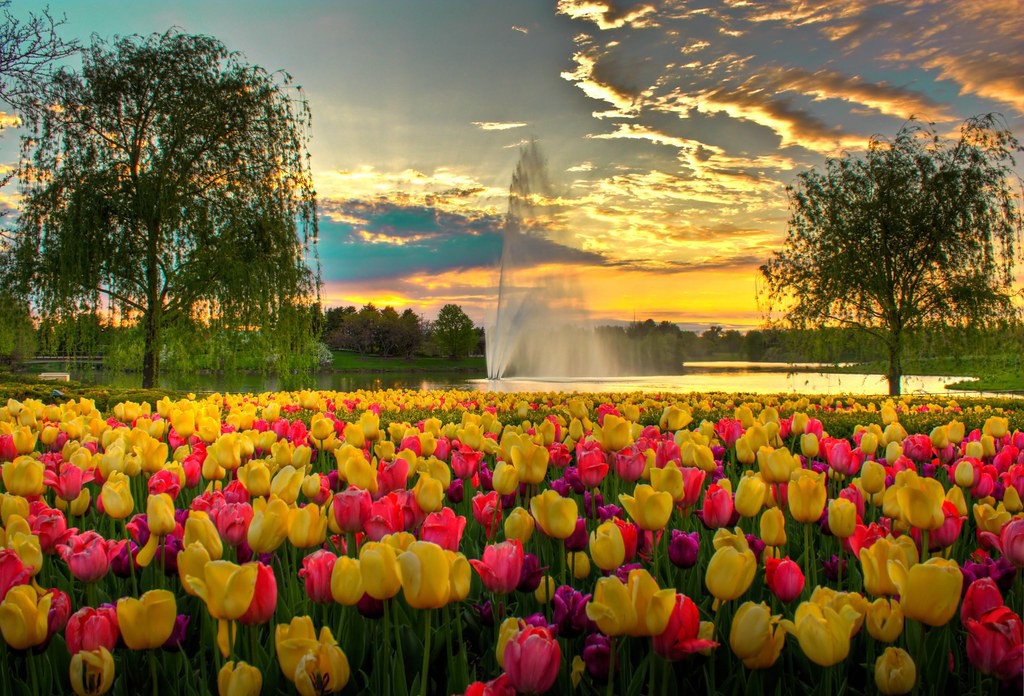}
        %\label{Fig: dims}
    \end{minipage}
   % \hspace{1mm}
    \hspace{-5mm}{+}
    \begin{minipage}{0.3\textwidth}
        \begin{abox} 
I am doing image-to-image translation. Could you think of creative prompts to translate this image to different ones? Keep them creative, and only return 20 different prompts.\par
\end{abox}
    \end{minipage}
    \vspace*{-2mm}
    \caption{The script for requesting GPT-4o to generate $20$ different prompts for each original image.} 
    \label{Fig: gpt}
    \vspace*{-7mm}
\end{figure}
As shown in Fig. \ref{Fig: gpt}, we request GPT-4o to generate $20$ different prompts for each original image.

\begin{table*}[t]
\caption{The influence of editing strengths on the performance of our method.} 
\vspace*{-2mm}
\small
\begin{tabularx}{\hsize}{|>{\raggedleft\arraybackslash}p{1.2cm}||Y|Y|Y|Y|Y|Y|Y|Y|Y|Y|Y|Y|Y|Y|}
    \hline\thickhline
    \rowcolor{mygray}
   &\multicolumn{2}{c|}{SD2 $\uparrow$} &\multicolumn{2}{c|}{SDXL $\uparrow$} &\multicolumn{2}{c|}{OpDa $\uparrow$} &\multicolumn{2}{c|}{CoXL $\uparrow$} &\multicolumn{2}{c|}{Kan3 $\uparrow$} &\multicolumn{2}{c|}{SD3 $\uparrow$} 
     & \multicolumn{2}{c|}{Kolor $\uparrow$} \\ 
   \rowcolor{mygray} \multirow{-2}{*}{Strength}& mAP & Acc & mAP & Acc & mAP & Acc & mAP & Acc & mAP & Acc & mAP & Acc & mAP & Acc \\ \hline \hline
   
    $0.1$ & $100.0$ & $100.0$ & $99.9$ & $99.9$ & $100.0$ & $100.0$ & $100.0$ & $99.9$ & $100.0$ & $100.0$ & $100.0$ & $100.0$ & $99.9$ & $99.9$ \\
    $0.2$ & $100.0$ & $100.0$ & $99.9$ & $99.9$ & $100.0$ & $100.0$ & $99.9$ & $99.9$ & $99.9$ & $99.9$ & $100.0$ & $99.9$ & $99.8$ & $99.8$ \\
    $0.3$ & $100.0$ & $100.0$ & $99.9$ & $99.8$ & $99.9$ & $99.9$ & $99.9$ & $99.8$ & $99.7$ & $99.7$ & $99.9$ & $99.9$ & $99.6$ & $99.5$ \\
    $0.4$ & $99.9$ & $99.9$ & $99.8$ & $99.7$ & $99.9$ & $99.8$ & $99.8$ & $99.7$ & $99.2$ & $99.1$ & $99.9$ & $99.8$ & $98.9$ & $98.6$ \\
    $0.5$ & $99.9$ & $99.8$ & $99.5$ & $99.3$ & $99.6$ & $99.4$ & $99.4$ & $99.2$ & $97.6$ & $97.1$ & $99.7$ & $99.6$ & $97.4$ & $96.9$ \\
    $0.6$ & $99.8$ & $99.7$ & $99.1$ & $99.0$ & $98.2$ & $97.8$ & $97.9$ & $97.5$ & $85.7$ & $83.3$ & $99.4$ & $99.2$ & $92.1$ & $90.8$ \\
    $0.7$ & $99.2$ & $99.0$ & $97.9$ & $97.5$ & $87.3$ & $85.3$ & $89.3$ & $87.7$ & $61.9$ & $57.2$ & $98.5$ & $98.1$ & $90.3$ & $88.8$ \\
    $0.8$ & $97.5$ & $97.1$ & $81.5$ & $78.8$ & $49.5$ & $45.4$ & $61.0$ & $57.1$ & $14.5$ & $11.4$ & $85.7$ & $82.9$ & $70.9$ & $67.4$ \\
    $0.9$ & $88.8$ & $86.6$ & $68.1$ & $63.9$ & $13.2$ & $10.7$ & $19.5$ & $16.5$ & $2.1$ & $1.5$ & $30.1$ & $25.6$ & $24.5$ & $20.5$ \\
    $1.0$ & $43.2$ & $37.7$ & $19.3$ & $15.7$ & $1.8$ & $1.2$ & $2.6$ & $1.8$ & $0.0$ & $0.0$ & $0.0$ & $0.0$ & $2.2$ & $1.5$ \\
    \hline
\end{tabularx}
\label{Table: editing}
\vspace*{-2mm}
\end{table*}

\begin{figure*}[t]
    \centering
    \hspace*{-2mm}
    \includegraphics[width=0.98\textwidth]{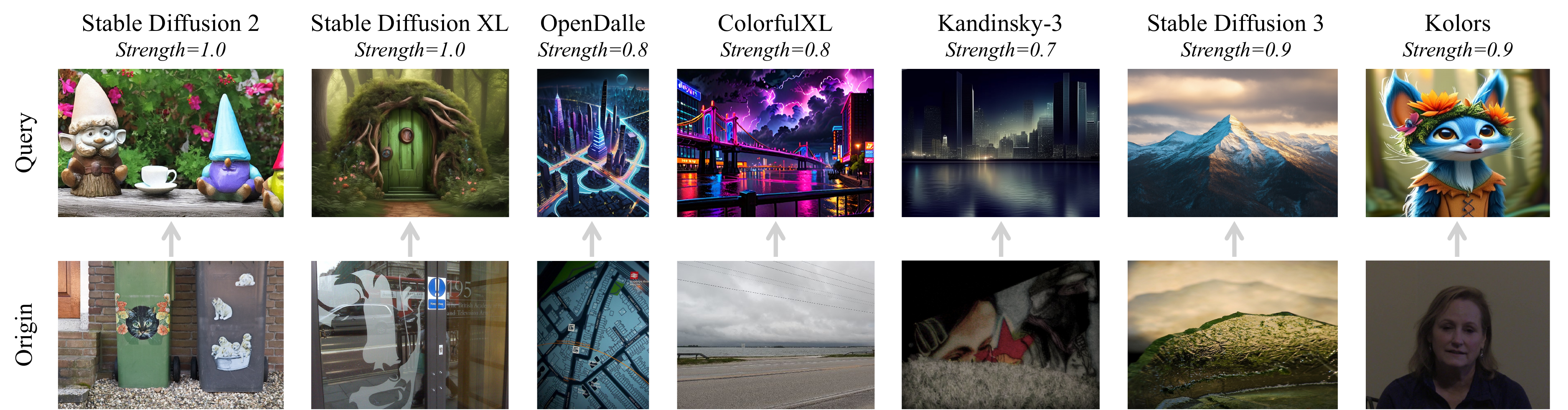}
    %\vspace{-1mm}
    \caption{Generations with large editing strengths: the queries are visually very dissimilar to the origins.} 
     \vspace{-4mm}
    \label{Fig: ff}
\end{figure*}

\section{Prompt and Generation Examples}\label{App: prompt}

In Fig. \ref{Fig: prompt}, we present several prompts with their corresponding generated images from our dataset, \dname~. The dataset comprehensively covers a wide range of subjects commonly found in real-world scenarios, such as natural sceneries, cultural architectures, lively animals, luxuriant plants, artistic paintings, and indoor items. It is important to note that in the training set, for each original image,  \dname~ contains 20 prompts with corresponding generated images, and for illustration, we only show 4 of them in Fig. \ref{Fig: prompt}.

\section{Influence of Editing Strengths}\label{App: Editing}
Table \ref{Table: editing} shows how the proposed method performs under different editing strengths. We observe that although the training and testing images come from different diffusion models with varying editing strengths, the performance of our method remains consistently high across most editing strengths. 
It is important to note that: (1) $strength = 1$ means it's almost like generating from pure noise, which is approximately equivalent to text-to-image generation. Therefore, it is reasonable that we cannot find the origins in that case; (2) Fig. \ref{Fig: ff} gives some examples of strengths where our method fails. These queries are indeed very visually dissimilar with the origins; and (3) We do not change the training editing strength for Stable Diffusion 2 while varying test editing strength. That means our method is also generalizable across varying editing strengths.

\begin{figure}[t]
    \centering
    \hspace*{-2mm}
    \includegraphics[width=0.47\textwidth]{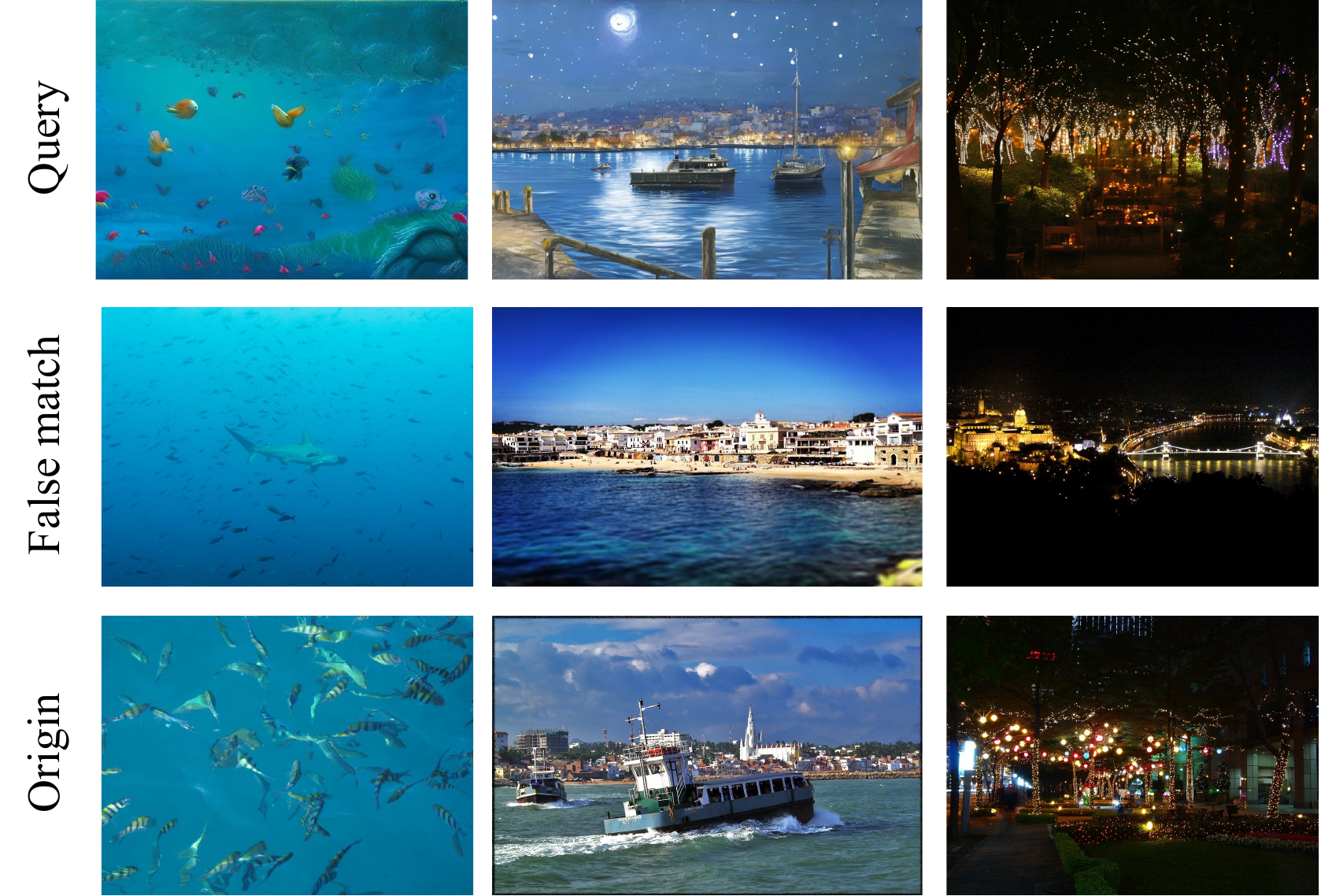}
    %\vspace{-1mm}
    \caption{This illustration shows failure cases predicted by our method. We have identified that our model may fail when encountering \textit{hard negative} samples.} 
     \vspace{-6mm}
    \label{Fig: failure}
\end{figure}

\begin{table*}[t]
\caption{The performance of publicly available models on 7 different diffusion models.} 
\vspace*{-2mm}
\small

\begin{tabularx}{\hsize}{|
>{\centering\arraybackslash}p{0.9cm}>{\raggedleft\arraybackslash}p{2cm}||Y|Y|Y|Y|Y|Y|Y|Y|Y|Y|Y|Y|Y|Y|}
    \hline\thickhline
    \rowcolor{mygray}
 &  &\multicolumn{2}{c|}{SD2 $\uparrow$} &\multicolumn{2}{c|}{SDXL $\uparrow$} &\multicolumn{2}{c|}{OpDa $\uparrow$} &\multicolumn{2}{c|}{CoXL $\uparrow$} &\multicolumn{2}{c|}{Kan3 $\uparrow$} &\multicolumn{2}{c|}{SD3 $\uparrow$} 
     & \multicolumn{2}{c|}{Kolor $\uparrow$} \\ 
   \rowcolor{mygray} &  \multirow{-2}{*}{Method}& mAP & Acc & mAP & Acc & mAP & Acc & mAP & Acc & mAP & Acc & mAP & Acc & mAP & Acc \\ \hline \hline
   
    \multirow{3}{*}{Supervised} & Swin-B & $3.1$ & $2.0$ & $2.9$ & $1.9$ & $4.1$ & $2.9$ & $4.2$ & $3.1$ & $6.8$ & $4.7$ & $2.9$ & $1.9$ & $3.0$ & $2.0$ \\
    \multirow{3}{*}{Pre-trained} & ResNet-50 & $3.8$ & $2.6$ & $3.1$ & $2.0$ & $5.3$ & $3.7$ & $4.5$ & $3.2$ & $8.1$ & $5.7$ & $3.4$ & $2.0$ & $3.4$ & $2.2$ \\
    \multirow{3}{*}{Models} & ConvNeXt & $3.5$ & $2.1$ & $3.3$ & $2.2$ & $4.7$ & $3.3$ & $5.0$ & $3.5$ & $8.4$ & $6.2$ & $3.5$ & $2.4$ & $3.6$ & $2.6$ \\
    & EfficientNet & $2.9$ & $1.9$ & $2.9$ & $2.0$ & $4.9$ & $3.4$ & $5.4$ & $3.9$ & $8.7$ & $6.5$ & $3.3$ & $2.2$ & $4.1$ & $3.0$ \\
    & ViT-B & $4.1$ & $2.8$ & $4.5$ & $3.1$ & $7.2$ & $5.5$ & $6.7$ & $5.0$ & $11.2$ & $8.7$ & $4.1$ & $2.8$ & $5.6$ & $4.3$ \\
    \hline\hline
    \multirow{2}{*}{Self-} & SimSiam & $1.5$ & $1.0$ & $1.2$ & $0.7$ & $1.8$ & $0.9$ & $1.7$ & $1.0$ & $3.1$ & $1.9$ & $1.5$ & $0.8$ & $1.4$ & $0.8$ \\
    \multirow{2}{*}{supervised}& MoCov3 & $1.4$ & $0.8$ & $1.5$ & $0.9$ & $2.4$ & $1.3$ & $2.2$ & $1.3$ & $3.8$ & $2.4$ & $1.9$ & $1.1$ & $1.6$ & $1.0$ \\
    \multirow{2}{*}{Learning}& DINOv2 & $2.6$ & $1.6$ & $2.7$ & $1.7$ & $4.6$ & $3.0$ & $5.5$ & $3.6$ & $8.4$ & $5.9$ & $2.9$ & $1.9$ & $3.6$ & $2.6$ \\
    \multirow{2}{*}{Models}& MAE & $14.9$ & $11.4$ & $10.0$ & $8.0$ & $13.1$ & $10.5$ & $8.1$ & $6.4$ & $17.6$ & $14.3$ & $11.2$ & $8.5$ & $6.5$ & $5.1$ \\
    & SimCLR & $6.0$ & $4.2$ & $7.0$ & $5.2$ & $13.5$ & $10.6$ & $13.0$ & $10.1$ & $23.7$ & $19.3$ & $7.3$ & $12.0$ & $8.8$ & $6.7$ \\
    \hline\hline
    \multirow{2}{*}{Vision-} & CLIP & $2.6$ & $1.7$ & $2.1$ & $1.4$ & $3.1$ & $2.1$ & $3.2$ & $2.0$ & $4.2$ & $2.7$ & $2.5$ & $1.6$ & $2.1$ & $0.7$ \\
    \multirow{2}{*}{language}& SLIP & $5.6$ & $3.8$ & $3.5$ & $2.3$ & $5.8$ & $4.0$ & $4.9$ & $3.3$ & $9.1$ & $6.7$ & $5.4$ & $3.5$ & $3.8$ & $2.5$ \\
    \multirow{2}{*}{Models}& ZeroVL & $5.2$ & $3.5$ & $4.4$ & $2.9$ & $6.4$ & $4.4$ & $4.5$ & $3.2$ & $9.8$ & $6.9$ & $4.7$ & $3.0$ & $4.3$ & $3.0$ \\
    & BLIP & $6.8$ & $4.8$ & $6.5$ & $4.5$ & $9.9$ & $7.0$ & $8.7$ & $6.3$ & $13.8$ & $10.2$ & $6.0$ & $3.9$ & $6.4$ & $4.5$ \\
    \hline\hline
    \multirow{3}{*}{Image Copy} & ASL & $2.3$ & $1.7$ & $3.0$ & $2.3$ & $5.6$ & $4.4$ & $5.7$ & $4.6$ & $10.3$ & $8.7$ & $2.7$ & $2.1$ & $3.7$ & $2.9$ \\
     \multirow{3}{*}{Detection}& CNNCL & $4.0$ & $2.9$ & $4.2$ & $3.2$ & $8.3$ & $6.7$ & $5.7$ & $4.5$ & $12.2$ & $9.9$ & $3.7$ & $2.7$ & $6.3$ & $5.0$ \\
    \multirow{3}{*}{Models}& BoT & $6.6$ & $4.9$ & $6.1$ & $4.4$ & $10.4$ & $8.2$ & $12.5$ & $10.2$ & $20.6$ & $16.8$ & $7.4$ & $5.4$ & $9.3$ & $7.3$ \\
    & SSCD & $9.7$ &$7.7$ & $8.7$ & $6.8$ & $16.4$ & $14.0$ & $18.1$ & $15.6$ & $28.1$ & $24.6$ & $9.0$ & $6.8$ & $14.1$ & $11.9$ \\
    & AnyPattern & $17.6$ & $14.3$ & $18.5$ & $15.7$ & $33.0$ & $29.2$ & $37.8$ & $34.0$ & $48.0$ & $43.9$ & $18.2$ & $15.0$ & $30.7$ & $27.5$ \\
    \hline
\end{tabularx}
\label{Table: all_1}
\vspace*{-1mm}

\caption{The performance of our trained models on 7 different diffusion models. Note that these models are trained on images generated by SD2 and tested on images from multiple models.} 
\vspace*{1mm}
\small
\begin{tabularx}{\hsize}{|
>{\centering\arraybackslash}p{0.9cm}>{\raggedleft\arraybackslash}p{2cm}||Y|Y|Y|Y|Y|Y|Y|Y|Y|Y|Y|Y|Y|Y|}
    \hline\thickhline
    \rowcolor{mygray}
 &  &\multicolumn{2}{c|}{SD2 $\uparrow$} &\multicolumn{2}{c|}{SDXL $\uparrow$} &\multicolumn{2}{c|}{OpDa $\uparrow$} &\multicolumn{2}{c|}{CoXL $\uparrow$} &\multicolumn{2}{c|}{Kan3 $\uparrow$} &\multicolumn{2}{c|}{SD3 $\uparrow$} 
     & \multicolumn{2}{c|}{Kolor $\uparrow$} \\ 
   \rowcolor{mygray} &  \multirow{-2}{*}{Method} & mAP & Acc & mAP & Acc & mAP & Acc & mAP & Acc & mAP & Acc & mAP & Acc & mAP & Acc \\ \hline \hline
   
    \multirow{1}{*}{Similarity-} & Circle loss & $70.4$ & $64.3$ & $56.2$ & $50.1$ & $56.5$ & $51.8$ & $41.6$ & $37.0$ & $60.0$ & $53.8$ & $65.6$ & $59.3$ & $43.5$ & $39.2$ \\ 
    \multirow{1}{*}{based}& SoftMax & $82.7$ & $78.3$ & $62.4$ & $56.5$ & $58.3$ & $53.0$ & $37.3$ & $32.2$ & $52.5$ & $46.0$ & $75.9$ & $70.2$ & $43.6$ & $38.7$ \\ 
    \multirow{1}{*}{Models}& CosFace & $87.1$ & $83.2$ & $63.7$ & $58.2$ & $56.7$ & $51.7$ & $30.5$ & $25.2$ & $47.5$ & $40.6$ & $71.5$ & $65.5$ & $43.0$ & $38.0$ \\ 
    \hline\hline
    \multirow{1}{*}{General-} & IBN-Net & $88.6$ & $85.1$ & $65.7$ & $60.1$ & $59.4$ & $54.2$ & $33.3$ & $28.3$ & $49.8$ & $42.8$ & $74.0$ & $68.3$ & $45.4$ & $40.5$ \\ 
    \multirow{1}{*}{izable}& TransMatcher & $65.6$ & $60.3$ & $60.6$ & $55.8$ & $67.9$ & $63.6$ & $61.7$ & $57.4$ & $68.9$ & $64.2$ & $64.7$ & $59.2$ & $67.9$ & $63.9$ \\ 
    \multirow{1}{*}{Models}& QAConv-GS & $78.8$ & $74.9$ & $71.6$ & $67.5$ & $77.4$ & $74.3$ & $73.6$ & $70.5$ & $75.2$ & $71.2$ & $77.3$ & $73.6$ & $79.5$ & $76.9$ \\ 
    \hline\hline
     & VAE Embed. & $51.0$ & $47.0$ & $38.3$ & $33.8$ & $42.3$ & $38.6$ & $51.6$ & $48.8$ & $54.7$ & $50.4$ & $47.7$ & $42.9$ & $46.9$ & $43.6$ \\ 
    \multirow{1}{*}{\textbf{Ours}}& Linear Trans. & $\hspace{-0.4mm}\mathbf{88.8}$ & $\hspace{-0.4mm}\mathbf{86.6}$ & $\hspace{-0.4mm}\mathbf{81.5}$ & $\hspace{-0.4mm}\mathbf{78.8}$ & $\hspace{-0.4mm}\mathbf{87.3}$ & $\hspace{-0.4mm}\mathbf{85.3}$ & $\hspace{-0.4mm}\mathbf{89.3}$ & $\hspace{-0.4mm}\mathbf{87.7}$ & $\hspace{-0.4mm}\mathbf{85.7}$ & $\hspace{-0.4mm}\mathbf{83.3}$ & $\hspace{-0.4mm}\mathbf{85.7}$ & $\hspace{-0.4mm}\mathbf{82.9}$ & $\hspace{-0.4mm}\mathbf{90.3}$ & $\hspace{-0.4mm}\mathbf{88.8}$ \\ 
    & Upper & $88.8$ & $86.6$ & $84.9$ & $82.4$ & $90.8$ & $89.2$ & $93.1$ & $91.9$ & $95.4$ & $94.3$ & $93.7$ & $92.0$ & $94.0$ & $92.8$ \\ 
    \hline
  \end{tabularx}
  \label{Table: all_2}
  \vspace*{-3mm}
\end{table*}

\begin{table*}[t]
  \caption{The generalization results on InstructPix2Pix \citep{brooks2023instructpix2pix}, IP-Adapter \citep{ye2023ip-adapter}, EDICT \citep{wallace2023edict}, and Plug-and-Play \citep{tumanyan2023plug}.} 
  \begin{tabularx}{\hsize}{|
>{\centering\arraybackslash}p{0.9cm}>{\raggedleft\arraybackslash}p{3cm}||Y|Y|Y|Y|Y|Y|Y|Y|Y|Y|Y|Y||}
    \hline\thickhline
    \rowcolor{mygray}
 &  &\multicolumn{2}{c|}{\scalebox{1}{InstructP2P $\uparrow$}} &\multicolumn{2}{c|}{\scalebox{1}{IP-Adapter $\uparrow$}} &\multicolumn{2}{c|}{\scalebox{1}{EDICT $\uparrow$}}&\multicolumn{2}{c|}{\scalebox{1}{Plug-and-Play $\uparrow$}} \\ 
   \rowcolor{mygray} &  \multirow{-2}{*}{Method}&  \scalebox{1}{mAP} & \scalebox{1}{Acc}&  \scalebox{1}{mAP} & \scalebox{1}{Acc}&  \scalebox{1}{mAP} & \scalebox{1}{Acc}&  \scalebox{1}{mAP} & \scalebox{1}{Acc}\\ \hline  
    
    \hline\hline
    & \scalebox{1}{VAE Embed.} 
    &$68.2$
    &$67.1$
    &$0.2$
    &$0.1$
    &$60.4$
    &$55.4$
    &$99.4$
    &$99.1$\\ 
     \multirow{-2}{*}{\textbf{Ours}}& \cellcolor{lightroyalblue}\scalebox{1}{Linear Trans. VAE}&\cellcolor{lightroyalblue}$80.7$
&\cellcolor{lightroyalblue}$79.2$
&\cellcolor{lightroyalblue}$0.4$
&\cellcolor{lightroyalblue}$0.2$
&\cellcolor{lightroyalblue}$89.0$
&\cellcolor{lightroyalblue}$86.6$
&\cellcolor{lightroyalblue}$99.8$
&\cellcolor{lightroyalblue}$99.7$\\ 
    \hline
  \end{tabularx}
  \label{Table: beyond}
  \vspace*{-4mm}

\end{table*}

%\vspace{2mm}
\section{Failure Cases and Potential Directions}\label{App: fail}

\textbf{Failure cases.} As shown in Fig. \ref{Fig: failure}, we observe that our model may fail when negative samples are too visually similar to the queries. This \textit{hard negative} problem is reasonable because
our model uses a VAE to compress high-dimensional inputs into a lower-dimensional latent representation. This compression tends to smooth out subtle local details, causing the model to lose critical fine-grained distinctions between similar yet different instances. Furthermore, the imposed prior encourages a uniform distribution in the latent space, forcing nuanced features from distinct instances into overlapping latent representations. These factors reduce the model’s capability to differentiate hard negatives from true positives.

\textbf{Potential directions.} The \textit{hard negative} problem has been studied in the Image Copy Detection (ICD) community, as exemplified by ASL \citep{wang2023benchmark}. It learns to assign a larger norm to the deep features of images that contain more content or information. However, this method cannot be directly used in our scenario because the query and reference here do not have a simple relationship in terms of information amount. Nevertheless, it offers a promising research direction from the perspective of information. Specifically, on one hand, the noise-adding and denoising processes result in a loss of information, while on the other hand, the text introduces new information into the output.

\vspace*{-1mm}
\section{Complete Experiments for 7 Models}\label{App: all}

We provide two types of complete experiments for seven different diffusion models: (1) Table \ref{Table: all_1} presents the results from directly testing publicly available models on the \dname~ test dataset; and (2) Table \ref{Table: all_2} shows the results from testing models that we trained on the \dname~ training dataset, which contains only images generated by Stable Diffusion 2.

%\begin{figure}[t]
%    \centering
 %   \hspace*{-2mm}
 %   \includegraphics[width=0.4\textwidth]{paradigm.pdf}
    %\vspace{-1mm}
 %   \caption{Different paradigms used by text-guided image-to-image translations.} 
  %   \vspace{-3mm}
  %  \label{Fig: paradigm}
%\end{figure}

\section{Limitations and Future Works}\label{App: future}

\textbf{Limitations.}
Although the paradigm analyzed in the main paper is the simplest approach for text-guided image-to-image translation and serves as the default mode in the $\mathtt{AutoPipelineForImage2Image}$ of $\mathtt{diffusers}$, we also observe the existence of alternative paradigms, such as InstructPix2Pix \citep{brooks2023instructpix2pix}, IP-Adapter \citep{ye2023ip-adapter}, EDICT \citep{wallace2023edict}, and Plug-and-Play \citep{tumanyan2023plug}. While these paradigms lie beyond our theoretical guarantees, we can still analyze them experimentally, as demonstrated in Table \ref{Table: beyond}. Interestingly, we find that \textbf{(1)} our method generalizes well to InstructPix2Pix, EDICT and Plug-and-Play, which still use a \textcolor{reda}{\textit{VAE encoder}} to embed the original images; and \textbf{(2)} we fail on IP-Adapter, which uses \textcolor{reda}{\textit{CLIP for encoding}}. We also try the linear transformed CLIP embedding, but it still fails to generalize ($36.6\%$ mAP and $27.8\%$ Acc). Based on these experiments, we conclude with a hypothesis about the upper limit of our method:

\tcbset{colback=lightroyalblue, colframe=white, left=1mm, right=1mm, top=1mm, bottom=1mm}
\vspace*{-1mm}
\begin{tcolorbox}[breakable]
\textbf{Hypothesis 1.} \textit{Following \textbf{Theorem 1}, consider a different well-trained diffusion model $\mathcal{F}_3$ and its text-guided image-to-image functionability achieved with \textcolor{reda}{\textbf{VAE-encoded original images}}. The matrix $\mathbf{W}$ can be generalized such that for any original image $o$ and its translation $g_3$, we have:}
\begin{equation}\label{Eq: H1}
\mathcal{E}_{1}(g_3) \cdot \mathbf{W} = \mathcal{E}_{1}(o) \cdot \mathbf{W}.
\end{equation}
\end{tcolorbox}
\vspace*{-1mm}

\textbf{Future Works.} Future works may focus on \textbf{(1)} providing a theoretical proof for Hypothesis 1, and \textbf{(2)} developing new generalization methods for text-guided image-to-image based on CLIP encodings.

\end{document}